\pdfoutput=1

\documentclass[11pt]{article}

\usepackage[final]{acl}

\usepackage{times}
\usepackage{latexsym}

\usepackage[T1]{fontenc}

\usepackage[utf8]{inputenc}

\usepackage{microtype}

\usepackage{inconsolata}

\usepackage{graphicx}

\usepackage{amsmath,amsfonts,bm}

\def\eqref#1{equation~\ref{#1}}

\def\1{\bm{1}}

\DeclareMathAlphabet{\mathsfit}{\encodingdefault}{\sfdefault}{m}{sl}
\SetMathAlphabet{\mathsfit}{bold}{\encodingdefault}{\sfdefault}{bx}{n}

\def\gC{{\mathcal{C}}}

\usepackage{wrapfig}
\usepackage{multirow,makecell}
\usepackage{tabularx}
\usepackage{subfigure}
\usepackage{bbding}
\usepackage{threeparttable}
\usepackage{wasysym}
\usepackage{todonotes}
\usepackage{enumitem}
\usepackage{makecell}

\usepackage{hyperref}       
\usepackage{url}            
\usepackage{booktabs}       
\usepackage{amsfonts}       
\usepackage{nicefrac}       
\usepackage{microtype}      
\usepackage{xcolor}         
\usepackage{pifont}
\usepackage{bm}
\usepackage{bbm}

\usepackage{color, colortbl}
\definecolor{Gray}{gray}{0.93}
\definecolor{Orange}{rgb}{1,0.5,0}
\definecolor{DGray}{gray}{0.83}
\definecolor{LightCyan}{rgb}{0.88,1,1}
\usepackage[most, breakable]{tcolorbox}

\usepackage{algorithm,algpseudocode}
\usepackage{amsthm}
\algnewcommand{\algorithmicforeach}{\textbf{for each}}
\algdef{SE}[FOR]{ForEach}{EndForEach}[1]
{\algorithmicforeach\ #1\ \algorithmicdo}
{\algorithmicend\ \algorithmicforeach}

\newcommand{\e}[1]{{\small $#1$}}

\newcommand{\alg}{\textsc{BTProp}}
\newcommand{\priorconf}{\textsc{Prior Confidence}}
\newcommand{\selfcheck}{\textsc{SelfCheckGPT}}
\newcommand{\maieutic}{\textsc{Maieutic Prompting}}
\newcommand{\semantic}{\textsc{Semantic Uncertainty}}
\newcommand{\focus}{\textsc{Focus}}
\newcommand{\chainofthought}{\textsc{CoT}}

\usepackage[most, breakable]{tcolorbox}
\usepackage{listings}
\lstdefinestyle{text}{
    basicstyle=\fontsize{8}{9}\ttfamily,
    showstringspaces=false,
    breaklines=true,
    breakatwhitespace=false,
    breakindent=0pt,
    keepspaces=true,
    showspaces=false,  
    escapeinside={(*@}{@*)},
}

\title{A Probabilistic Framework for LLM Hallucination Detection \\via Belief Tree Propagation}

\author{
  \textbf{Bairu Hou}\textsuperscript{1}\thanks{Correspondence to: Bairu Hou
$<$bairu@ucsb.edu$>$.}\quad
  \textbf{Yang Zhang}\textsuperscript{2}\quad
  \textbf{Jacob Andreas}\textsuperscript{3}\quad
  \textbf{Shiyu Chang}\textsuperscript{1}\quad
\\
\\
  \textsuperscript{1}UC Santa Barbara\quad\quad
  \textsuperscript{2}MIT-IBM Watson AI Lab\quad\quad
  \textsuperscript{3}MIT CSAIL
}

\begin{document}
\maketitle
\begin{abstract}
We describe Belief Tree \textsc{Prop}agation (\alg), a probabilistic framework for LLM hallucination detection. To judge the truth of a statement, \alg~generates a \emph{belief tree} by recursively expanding the initial statement into a set of logically related claims, then reasoning globally about the relationships between these claims. \alg~works by constructing a probabilistic model of the LM itself: it reasons jointly about logical relationships between claims and relationships between claim probabilities and LM factuality judgments via probabilistic inference in a ``hidden Markov tree''.
This method improves over state-of-the-art baselines by 3\%-9\% (evaluated by AUROC and AUC-PR) on multiple hallucination detection benchmarks. Code is available at \url{https://github.com/UCSB-NLP-Chang/BTProp}.
\end{abstract}

\section{Introduction}

Current large language models (LLMs) often produce factually incorrect statements (sometimes referred to as ``hallucinations''). One popular approach to detecting hallucinations is to prompt LLMs to assign probabilities to the correctness of their own outputs (sometimes referred to as model ``beliefs"; \citealp{mitchell2022enhancing,hase2023methods, kassner2023language}). However, these beliefs are themselves frequently incorrect, miscalibrated, or inconsistent with each other.
To improve the accuracy of LM factuality judgments, a large body of work adjusts the assessed probability of a target claim by reasoning about sets of related statements \cite{
manakul2023selfcheckgpt,akyurek2024deductive, cao2023autohall, mundler2023self}.
These approaches typically involve two steps: first, the LLM is prompted to generate an augmented set of statements that support or contradict the target statement, possibly with probabilities attached \cite{jung2022maieutic%
};
second, these claims are combined via logical or probabilistic inference. As a simple example, if an LM assigns high probability to a target statement, but even higher probability to a contradictory statement, it may be reasonably inferred that the target statement is actually incorrect.

\begin{figure}
    \centering
    \includegraphics[width=0.48\textwidth]{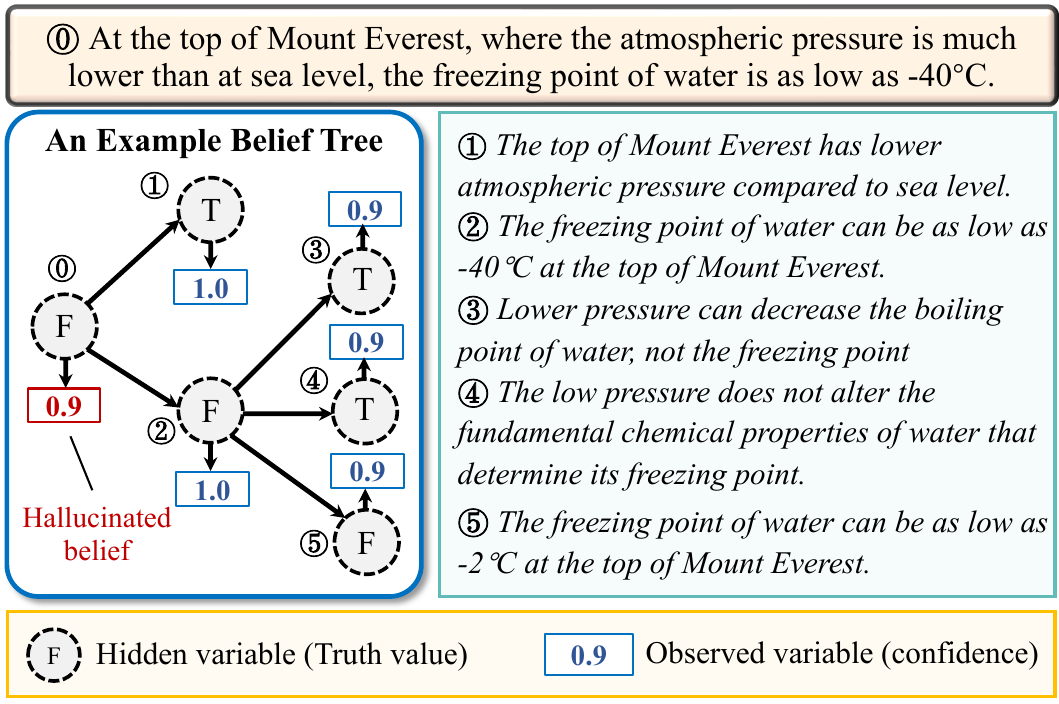}
    \caption{An example constructed belief tree.} 
    \label{fig: belief_tree_intro}
    \vspace{-5mm}
\end{figure}

One key limitation of current methods is that they do not model noise or uncertainty in the LM generation process itself. LLMs' ``beliefs'' are often poorly calibrated -- even statements that models judge to be true with 100\% probability are sometimes false. How can we produce reliable judgments about sets of related statements when even their assessments of individual statements are unreliable? 

To do so, we propose Belief Tree \textsc{Prop}agation (\alg), a probabilistic method for improving the accuracy LLM ``beliefs'' with applications to hallucination detection.
Specifically, given an initial statement generated by an LLM,  {\alg} first generates a \emph{belief tree} of recursively generated claims that are logically related ot the initial statement.
An example is shown in Figure~\ref{fig: belief_tree_intro}, in which the root node of the belief tree is one sentence generated by the LLM about the freezing point of 
water, and
each child node supports or contradicts its parent.
Next, {\alg} interprets this belief tree as a directed graphical model (a ``hidden Markov tree"; \citealp{crouse1998wavelet,durand2004computational}) in which the LLM's confidence scores are modeled as observed variables, and the ground-truth truthfulness of the statements as the corresponding hidden variables. 
Thus, the problem of assigning probabilities to statements can be reduced to probabilistic inference in a well-studied model family. Here, inference makes it possible to integrate knowledge about the (logical) structure of related statements, and the (probabilistic) relationship between statements and noisy LM outputs.
Experiment results show that our method achieves state-of-the-art performance, improving the hallucination detection performance (AUROC and AUC-PR) by 3\%--9\% on \texttt{FELM-Science}~\cite{chen2023felm} and \texttt{FactCheckGPT}~\cite{fadeeva2024fact} datasets relative to state-of-the-art baselines.

\section{Related Work}
\paragraph{Hallucinations in LLMs.}
Hallucination has become a prominent topic in the research of LLMs~\cite{liu2022token, dhuliawala2023chain, azaria2023internal, li2023halueval, zhang2023language}. 
Prior work~\cite{ji2023survey, huang2023survey} characterizes hallucination into two main types: factuality hallucination and faithfulness hallucination. Factuality hallucinations refer to outputs that are inconsistent with real-world facts. In comparison, faithfulness hallucinations shift to the generated content that deviates from the user's instruction or user-provided contextual information. In this paper, we mainly focuses on factuality hallucinations and aim to better leverage the LLM's intrinsic capabilities to detect such nonfactual statements within their generations.

\paragraph{Hallucination detection.} 
To detect hallucinations in LLM generation, the retrieval-based methods~\cite{shuster2021retrieval,min2023factscore, chern2023factool, semnani2023wikichat, huo2023retrieving, zhang2024knowhalu} compare the LLM's output with information from a reliable knowledge base. 
When an external knowledge base is inaccessible, another category of methods rely on the model's intrinsic capabilities. Among them, some approaches leverage the reasoning ability of LLMs to detect and reduce hallucinations~\cite{chen2023felm, dhuliawala2023chain} via chain-of-thought prompting~\cite{wei2022chain}. 
Uncertainty-based methods perform token-level or sentence-level uncertainty quantification~\cite{varshney2023stitch, fadeeva2024fact} to predict the factuality of model's generation.
Probing the model's  hidden states can also help detect hallucinations~\cite{azaria2023internal, chen2023inside, su2024unsupervised}.
The sampled-based methods (or consistency-based methods)~\cite{manakul2023selfcheckgpt, cao2023autohall, mundler2023self} sample multiple responses and then perform consistency checks between these responses and the original model generation. Inconsistencies are then used to detect hallucinations. Our method is closely related to the consistency-based methods. Instead of sampling additional responses and performing unstructured consistency checks, our method generates logically-related statements organized in a tree structure. We further propose a probabilistic framework based on the hidden Markov tree model to check consistency and detect hallucinations.

\paragraph{Improving factuality via logical consistency.} A growing body of research explores improving LLMs' factuality using the logical consistency across their beliefs~\cite{dalvi2021explaining, tafjord2022entailer, mitchell2022enhancing, jung2022maieutic, hase2023methods, kassner2023language, akyurek2024deductive}. Starting from an initial text (\emph{e.g.}, a statement requiring truthfulness assessment or a question), these methods identify additional texts that are logically connected to the initial text and organize them in tree or graph structures. The inconsistencies of model's beliefs on the factual correctness of these texts are resolved based on the inferential relations among these texts.
One main limitation is that they treat the beliefs (LM-generated truth values or probabilities) as calibrated, which is generally not the case. 
Motivated by this, we build a probabilistic framework to integrate the model's beliefs and figure out the most possible and reliable correctness evaluation of the initial text.

\section{Method}

\label{subsec:formulation}

Given a statement $v$ from the LLM-generated text, \emph{e.g.}, \emph{A star's temperature is determined by the amount of mass and energy it has}, and an LLM, we aim to use the LLM to gauge the correctness of the statement.
Many previous approaches use simple prompting techniques to directly elicit the LLM's ``belief'', or assessed probability of the statement, as a \emph{confidence score} in $[0, 1]$. For example, one approach is to directly query the LLM for whether this is true. By sampling multiple answers from the LLM, we can obtain the confidence score as the fraction of answers that say \emph{yes} (or some paraphrase thereof).
Crucially, these scores may be wrong in several ways: in addition to being factually incorrect, they may be \emph{miscalibrated} (reflecting an inappropriate degree of certainty or uncertainty) or \emph{inconsistent} (e.g., assigning high probability to mutually incompatible statements, or different probabilities to logically equivalent statements). Below, we describe a method that addresses both categories of error by reasoning jointly about relationships between the truth values of related statements and LLM confidence scores.

\subsection{\alg: An Intuitive Overview}
\label{subsec:overview}

The basic idea of our approach is to construct a \emph{belief tree}, denoted \e{\mathcal{T}}, in which the root node is the target statement, each child node is a statement logically related to the parent node, and each edge represents the logical relationship between two nodes. We then obtain the confidence scores of all the nodes and use the logical consistency among them to correct any potential mistakes in the scores.

\begin{figure}
    \centering
    \includegraphics[width=0.98\linewidth]{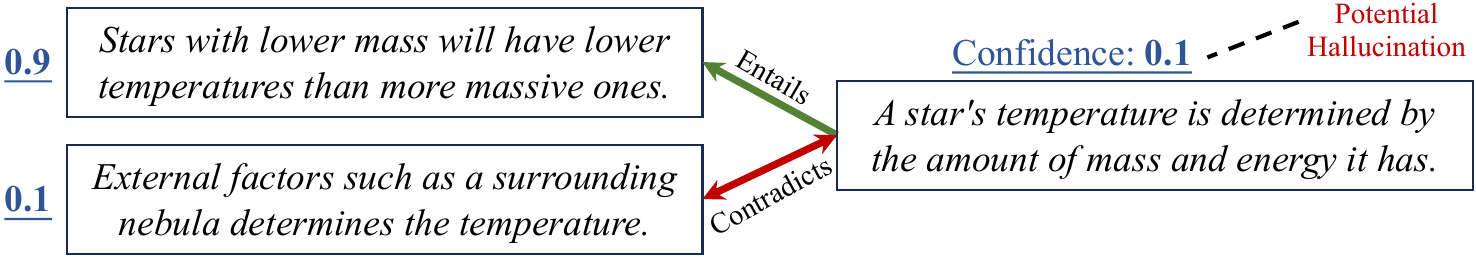}
    \caption{Motivating example for the proposed method.} 
    \vspace{-4mm}
\label{fig: motivating_example}
\end{figure}

Figure~\ref{fig: motivating_example} shows an example, where the target statement is \emph{A star's temperature is determined by the amount of mass and energy it has}. This statement is true but assume that the LLM produces a low confidence score, $0.1$, for the statement.
To correct this error, we can construct a belief tree \e{\mathcal{T}} by generating two child nodes from the target node, where the first, which is true, is entailed by the target statement and the second, which is false, contradicts the target statement. Assume that the LLM correctly assigns a high confidence score to the former and a low score to the latter. In this case, we can easily recognize that the belief of the target node is logically inconsistent with those of the child nodes, and that increasing the confidence score of the target node would resolve the logical inconsistency, thereby correcting the mistake.

The example above shows a simple case in which LLM beliefs about child nodes are all correct. In reality, these child nodes' beliefs may themselves be incorrect. To mitigate these additional errors, we can construct a much larger belief tree with greater depth and many different statements. Assuming the errors in the confidence scores are sporadic, we can then expect to correct most of these errors, in both the target and child nodes, by taking into account all the confidence scores and their logical consistency. An example of such a full belief tree is shown in Figure~\ref{fig: belief_tree_example}.

As implied by the discussion above, our algorithm consists of two components: \ding{182} Constructing the belief tree, and \ding{183} inferring the truthfulness of the target statement by integrating the confidence scores across the belief tree. Section~\ref{subsec:belief_tree_construct} will discuss the former and Sections~\ref{subsec:THMM}-\ref{subsec:inference} the latter.

\subsection{Belief Tree Construction}
\label{subsec:belief_tree_construct}

Since the belief tree \e{\mathcal{T}} consists of nodes as statements and edges as the logical relationships between parent and child statements, the construction of the belief tree iterates between the following two steps. \textbf{Step 1:} Given a statement as the parent node, generate a set of logically connected statements as the children nodes. \textbf{Step 2:} Determine the logical connection between the children nodes and parent node. The iterative process starts with the target statement as the root node, and terminates when the maximum tree depth is reached. Each step is detailed below.

\paragraph{Generating Child Statements}
As will be described below, \alg{} is general, and compatible with any method for generating belief tree nodes from parent statements. We explore three specific strategies for statement generation:

\textbf{\textit{Strategy 1: Statement Decomposition.}} For statements containing multiple facts/claims, we decompose them into individual sub-statements. Figure~\ref{fig: belief_tree_intro} shows an example where the target statement (\emph{Node 0}) contains multiple facts (the atmosphere pressure level and the freezing point of water on Mound Everest), It is then decomposed into two statements, one on the atmosphere pressure level and one on the freezing point of water. Statement decomposition can be achieved by prompting the LLM with in-context examples, as listed in Appendix~\ref{appendix: prompt}.

\textbf{\textit{Strategy 2: Supportive and Contradictory Premises.}} We prompt the LLM to generate a set of premises that are supportive or contradictory to the parent claim. When verifying the truthfulness of \emph{Node 2} in Figure~\ref{fig: belief_tree_intro} about the freezing point of water, we can prompt the model to generate contradictory premises (\emph{Node 3} and \emph{Node 4}) that implies the limited influence of low pressure on the freezing point of water. By leveraging these generated premises, we can indeed correct the model's wrong belief on \emph{Node 2} since the model's confidence scores on these generated premises are high in this example.

\textbf{\textit{Strategy 3: Statement Correction.}} In this strategy, the LLM is instructed to generate what it believes to be corrected versions of the parent statement. The \emph{Node 5} in Figure~\ref{fig: belief_tree_intro} shows one example of the statement correction process, where the statement is about the freezing point of water. Then the corrected statements would be almost the same as the parent statement (\emph{Node 2}), except that the actual temperature is replaced with alternatives that LLM believes may be true. This strategy is implemented via a three-step prompting process.
First, the LLM is prompted to generate a question about the key information in the statement (\emph{e.g.}, the freezing point in \emph{Node 2} in Figure~\ref{fig: belief_tree_intro}). Then, the we sample answers from the LLM. Finally, the LLM is prompted to generate a corrected statement from the original statement if it is wrong according to each sampled answer.
Note that there can be multiple corrected statements because the LLM may produce different answers when sampled multiple times. Also, the corrected statements may include the parent statement itself if the LLM believes the parent statement is a likely answer. More details about this process can be found in Appendix~\ref{appendix: prompt}.

Our method select the most appropriate strategies for each parent claim as follows. First, LLM first attempt the statement decomposition strategy. If the decomposition returns multiple statements, which indicates that the decomposition is meaningful, we would use them as the child statements. If only a single statement is returned, indicating that the parent statement cannot be further decomposed, we will prompt the LLM to select between strategies 2 and 3 with a list of rules and examples. The detailed prompts are  in Appendix~\ref{implement_detail}.

\paragraph{Determining Logical Relationships} Given a pair of parent and child statements, $u$ and $v$, we consider the following four logical relationships: \ding{182} Equivalence, $u \Leftrightarrow v$, if $u$ entails $v$ and $v$ entails $u$; \ding{183} entailment, $u \Rightarrow v$, if $u$ entails $v$ but $v$ is neutral to $u$; \ding{184} reverse entailment, $u \Leftarrow v$, if $u$ is neutral to $v$ but $v$ entails $u$; and \ding{185} contradiction, $u \Rightarrow \neg v$ (or, equivalently, $v \Rightarrow \neg u$).
Note that we do not consider the completely neutral relationship because any statements determined as completely neutral to their parent statement will be removed.

For the decomposition strategy (strategy 1), the child node statements are, by construction, \emph{jointly equivalent to} the parent statement. Formally, denote $u$ as the parent node and \e{\mathcal{C}(u)} as the set of its child nodes, then their logical relationship is determined as $u \Leftrightarrow \cap_{v \in \mathcal{C}(u)} v$.

For strategies 2 and 3, since each child statement is independently generated, we only need to determine their \emph{individual} relationship to the parent statement, rather than the joint relationship.
To this end, we leverage an off-the-shelf natural language inference (NLI) model to infer the entailment, neutrality, or contradiction between the statements. For each pair of parent statement $u$ and an \emph{individual} child statement $v$, we derive two NLI relationships, one by setting $u$ as the premise and $v$ as the hypothesis, and the other with $u$ and $v$ switched. The two NLI outputs are then mapped to the aforementioned logical relationships -- \textit{(entail, entail)} mapped to \textit{equivalence}; \textit{(entail, neutral)} mapped to \textit{entailment}; \textit{(neutral, entail)} mapped to \textit{reverse entailment}; and any results containing \textit{contradict} in either direction mapped to \textit{contradiction}. 
If the NLI module returns \textit{(neutral, neutral)}, the corresponding child statement will be discarded.

\paragraph{Prior Belief Estimation} The last step of the belief tree construction is to estimate the model's belief (\emph{i.e.}, whether the statement is true) on each node. 
Specifically, we directly probe the LLM with the prompt `\texttt{True or False? \{target statement\}}' and use the next token prediction probabilities of the words \textit{`True'} and \textit{`False'} to compute the model confidence. 
We normalize the prediction logits of the two words to get the confidence score of that statement. The only exception is in the case of statement correction, where the LLM is already instructed to output alternative statements it believes to be true. As a result, we simply set the confidence score of each generated statement to 1. Our empirical analysis shows that this would greatly reduce the number of LLM queries without deteriorating the performance.

\subsection{A Hidden Markov Tree Model}
\label{subsec:THMM}

After the belief tree is constructed, the next question is how to utilize the confidence scores across the tree to better determine the truthfulness of the root statement, namely the target statement. To this end, we introduce a hidden Markov tree model and frame the truthfulness estimation problem as a probabilistic inference problem.

\begin{figure}
    \centering
    \includegraphics[width=0.48\textwidth]{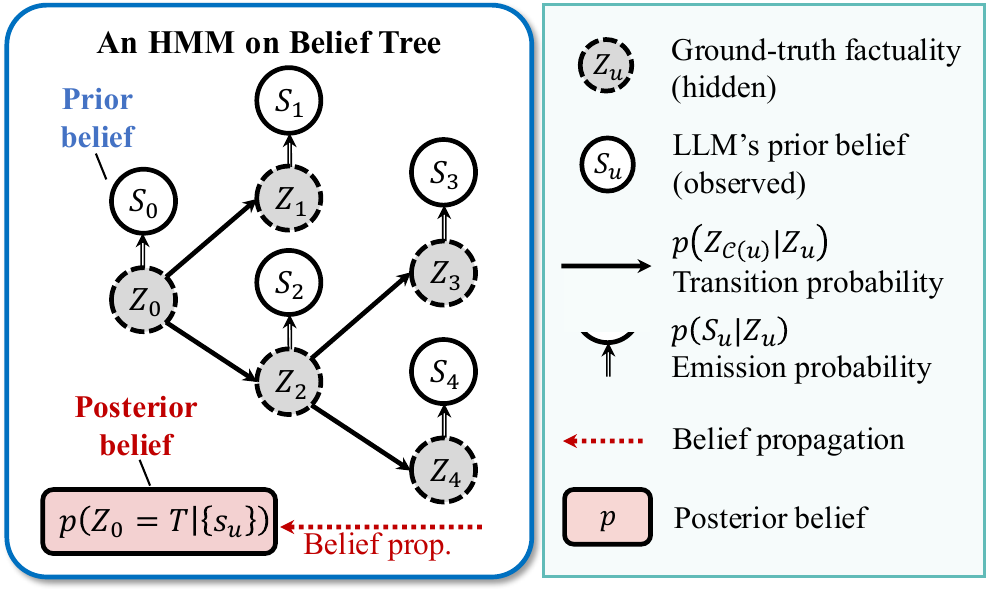}
    \vspace{-5mm}
    \caption{An example hidden Markov tree model.} 
    \label{fig: belief_tree_example}
    \vspace{-3mm}
\end{figure}

Figure~\ref{fig: belief_tree_example} shows an example hidden Markov tree model built upon the belief tree, where there are two layers of variables.
The upper layer consists of the confidence score of each statement, denoted \e{\{S_u\}}, which is estimated during the belief tree construction process.
The lower layer consists of the binary variables representing the actual correctness of each statement, denoted as \e{\{Z_u\}}. \e{Z_u = T} if statement $v$ is correct and \e{Z_u = F} otherwise. \e{\{S_u\}} are observed variables while \e{\{Z_u\}} are hidden variables. 

Given the above hidden Markov tree model, determining the truthfulness of the target statement can be cast as computing the posterior probability, \emph{i.e.}, \e{p(Z_0 = T \mid \{S_u\})}, which means the truthfulness of the root node given the confidence scores on all the nodes in this belief tree.
To computing \e{p(Z_0 = T \mid \{S_u\})}, we need to estimate the following probabilities.

The first set is the \emph{emission probability}, \e{p(S_u \mid Z_u)}, which characterizes the LLM's confidence score of the statement given its underlying truthfulness. The emission probability captures the quality of LLM's original confidence score. If the confidence score is accurate, then \e{p(S_u \mid Z_u =T)} will concentrate its probability mass towards $1$, and \e{p(S_u \mid Z_u =F)} will concentrate its probability mass towards $0$. Otherwise, the conditional distributions will be more spread.

The second set of distribution is the \emph{transition probability}, \e{p(Z_{\mathcal{C}(u)} \mid Z_{u})}, where \e{Z_{\mathcal{C}(u)}} represent the set of truthfulness variables of all the child nodes of $u$. The transition probability is the joint probability of the truthfulness of the child statements given that of the parent statement, which captures the logical relationships among these statements.

The third set of distribution is the \emph{prior distribution} of hidden variables, \emph{i.e.}, \e{p(Z_u = z)}, which refers to the probability of an arbitrary statement $u$ generated by the LLM being true or false. We use uniform prior (both 0.5 for $z=T$ and $z=F$) in this paper.

Therefore, the process of computing the posterior probability contains two steps: \ding{183} the estimation step, to first estimate these probabilities on a held-out dataset. \ding{184} the inference step, to compute the posterior probability for each testing example. Section~\ref{subsec:estimation} will discuss how to determine the emission and transition probabilities and Section~\ref{subsec:inference} how to infer \e{p(Z_0 = T \mid \{S_u\})}.

\subsection{Determining Conditional Probabilities}
\label{subsec:estimation}

\paragraph{Determining Emission Probabilities} The emission probabilities, \e{p(S_u \mid Z_u)}, can be estimated from labeled datasets. Specifically, given a dataset of statements with truthfulness labels, we can run the LLM to obtain the confidence scores of all the statements. We can then estimate \e{p(S_u \mid Z_u=T)} by obtaining the empirical distribution of the confidence scores within the statements labeled as true, and \e{p(S_u \mid Z_u=F)} within the statements labeled as false. To obtain the empirical distribution of the continuous confidence scores, we first quantize the support \e{[0, 1]} into multiple bins, and count the histogram of the confidence scores falling into each bin. The boundaries of each bin are listed in Table~\ref{tab: emission probability}. Given a particular confidence score on a statement, we look up the table to get the corresponding emission probability. For example, a confidence score 0.95 leads to \e{p(S_u = 0.95 \mid Z_u = F) = 0.65}.

\begin{table}
    \centering
    \resizebox{0.95\linewidth}{!}{
    \begin{tabular}{l|ccccc}
    \toprule[1pt]
    & $S_u\in [0.0, 0.2)$ & $[0.2, 0.4)$ & $[0.4, 0.7)$ & $[0.7, 0.9)$ & $ [0.9, 1.0]$ \\
    \midrule
    $Z_u = \texttt{True}$ & 0.12 & 0.05 & 0.10 & 0.08 &  0.65 \\
    $Z_u = \texttt{False}$ & 0.30 & 0.10 & 0.15 & 0.13 & 0.32\\
\bottomrule[1pt]
    \end{tabular}
    }
    \vspace{-2mm}
    \caption{Empirical estimation of the emission probability on \texttt{Wikibio-GPT3} dataset~\cite{manakul2023selfcheckgpt}.}
    \label{tab: emission probability}
    \vspace{-3mm}
\end{table}

\paragraph{Determining Transition Probabilities}

The transition probability, \e{p(Z_{\mathcal{C}(u)} \mid Z_{u})}, is a multivariate Bernoulli distribution with the support \e{\{T, F\}^m}, where $m$ is the size of \e{\mathcal{C}(u)}. As discussed in Section~\ref{subsec:belief_tree_construct}, different statement generation strategies have different logical relationship properties. For the statement decomposition strategy, the child statements are always jointly equivalent to the parent statement. 
Therefore, when the parent claim is true, \emph{i.e.}, \e{Z_{u}=T}, we know that all of the child statements must be true, and thus \e{p(Z_{\mathcal{C}(u)} = \bm z \mid Z_{u} = T)} equals one when every element in $\bm z$ is true and zero otherwise. Conversely, when the parent claim is false, \emph{i.e.}, \e{Z_u = F}, we can infer that at least one child statement is wrong, so \e{p(Z_{\mathcal{C}(u)} = \bm z \mid Z_{u} = F)} is set to zero when every element in $\bm z$ is true and uniformly set to \e{1/(2^m - 1)} otherwise.

For the other two strategies, the child statements of the same parent statement are considered independent conditional on the parent statement. Therefore, we can decompose the probability \e{p(Z_{\mathcal{C}(u)} \mid Z_{u})} into \e{\prod_{v \in \mathcal{C}(u)} p(Z_v \mid Z_u)}, where each \e{p(Z_v \mid Z_u)} is the transitional probability between parent statement $u$ and one child statement $v$. \e{p(Z_v \mid Z_u)} is determined based on the four different possible logical relationships between $u$ and $v$ (recall the discussion in Section~\ref{subsec:belief_tree_construct}), which is listed in Table~\ref{tab: transition_prob}.

\begin{table}[t]
    \centering
    \resizebox{\linewidth}{!}{
    \begin{tabular}{c|cc|cc|cc|cc}
    \toprule[1pt]
    & \multicolumn{2}{c|}{$u \Leftrightarrow v$} 
    & \multicolumn{2}{c|}{$u \Rightarrow v$} 
    & \multicolumn{2}{c|}{$u \Leftarrow v$} 
    & \multicolumn{2}{c}{$u \Rightarrow \neg v$} \\ 
    \midrule
    & $u = T$ & $u=F$
    & $u = T$ & $u=F$
    & $u = T$ & $u=F$
    & $u = T$ & $u=F$\\
    $v = T$ & 1.0  & 0.0 & 1.0  & $p_t$ & $p_t$ & 0.0 & 0.0  & $p_t$\\
    $v = F$ & 0.0 & 1.0 & 0.0 & $p_f$ & $p_f$ & 1.0 & 1.0 & $p_f$ \\
    \bottomrule[1pt]
    \end{tabular}
    }
    \vspace{-2mm}
    \caption{Transition probability given different logical relationships between two nodes. $p_t$ and $p_f$ are set to $0.5$ in this paper.}
    \label{tab: transition_prob}
    \vspace{-3mm}
\end{table}

\subsection{Inferring the Posterior Probability of $Z_0$}
\label{subsec:inference}

Inferring the posterior probability \e{p(Z_0 = T \mid \{S_u\})} is a standard inference problem in hidden Markov tree models and we can apply the standard \emph{upward-downward algorithm}~\cite{crouse1998wavelet} to efficiently compute this probability. For brevity, we only describe the gist of the algorithm here.
More details can be found in Appendix~\ref{appendix: detailed_derivation} and previous work~\cite{crouse1998wavelet, durand2004computational}.

The algorithm introduces an auxiliary conditional distribution \e{\beta(z, u)\equiv p(S_{\mathcal{T}(u)} \mid Z_u = z )}, where \e{z \in \{T, F\}} and \e{S_{\mathcal{T}(u)}} represents the confidence scores of all the nodes in the sub-tree whose root node is $u$. Note that the posterior probability of our interest, \e{p(Z_0 = T \mid \{S_{\mathcal{T}(0)}\})}, can be computed as

\vspace{-3mm}
{\small
\begin{equation*}
    \begin{aligned}
        p(Z_0 = T \mid \{S_{\mathcal{T}(0)}\}) = \frac{\beta(T, 0) p(Z_0 = T)}{\sum_{z\in\{T,F\}}\beta(z, 0) p(Z_0 = z)},
    \end{aligned}
\end{equation*}
}
\vspace{-3mm}

\noindent where \e{p(Z_0 = T)} and \e{p(Z_0 = F)} are exactly the prior probabilities we mentioned in Section~\ref{subsec:THMM}, and both are set to be 0.5.
Therefore, the posterior probability above can be represented as \e{\beta(T,0) / (\beta( F, 0) + \beta(T, 0))}.
According to the Bayesian rule, we can derive a recursive relationship between \e{\beta(z, u)} of a parent node $u$ and those of the child nodes:
\vspace{-3mm}

{\small
\begin{equation}
\begin{aligned}
    &\beta(z, u) = p(S_u \mid Z_u = z) \cdot \\
    &\sum_{Z_{\mathcal{C}(u)} \in \{T, F\}^m}p(Z_{\mathcal{C}(u)} \mid Z_u = z) \prod_{v \in \mathcal{C}(u)} \beta(Z_v, v),   
\end{aligned}
    \label{eq:ud}
\end{equation}
}
\vspace{-3mm}

\noindent where $m$ is the size of \e{\mathcal{C}(u)} (namely the number of child statements to node $u$). Note that the first term on the RHS is the emission probability, and the first term inside the summation is the transition probability, which are both already known. Therefore, Equation~\ref{eq:ud} provides a way to compute \e{\beta(z, 0)} recursively from the leaf nodes back to the root node.
First, we compute the \e{\beta(z, u)} of the leaf nodes as \e{\beta(z, u) = p(S_u \mid Z_u = z)}.
Then, we use Equation~\ref{eq:ud} to recursively compute the \e{\beta(z, u)} of a parent node $u$ from their child nodes, until the root node is reached. The entire process is essentially propagating and merging the beliefs in the confidence scores from sub-trees upward to the parent node. By the time we reach the root node, we have gathered the information of all the confidence scores. Hence this process is also referred to as a \emph{belief propagation} process.

We summarize the whole algorithm in Appendix~\ref{subsec: algrithm_table}.

\section{Experiments}
In this section, we conduct empirical evaluations on widely-used hallucination detection benchmarks.

\subsection{Experiment Configurations}
\label{subsec: exp_config}
\paragraph{Datasets} We follow the previous work of hallucination detection~\cite{manakul2023selfcheckgpt, chen2023felm} and use the following datasets for evaluation: \texttt{Wikibio-GPT3}~\cite{manakul2023selfcheckgpt}, \texttt{FELM-Science}~\cite{chen2023felm}, and \texttt{FactCheckGPT}~\cite{wang2023factcheck}. \texttt{Wikibio-GPT3} mainly consists of biography articles generated by LLMs,
whereas the other two datasets cover a broader range of topics such as physics, chemistry, and computer science.

\newcommand{\orangecolor}[1]{\textcolor{lightorange}{#1}}
\newcommand{\gr}{\rowcolor[gray]{.95}}
\definecolor{rulecolor}{RGB}{0,71,171}
\definecolor{tableheadcolor}{RGB}{204,229,255}
\begin{table*}[t]
\centering
\definecolor{rulecolor}{RGB}{0,71,171}
\definecolor{tableheadcolor}{RGB}{204,229,255}
\newcommand{\myrowcolour}{\rowcolor{tableheadcolor}}
\newcommand{\highest}[1]{\textcolor{blue}{\textbf{#1}}}
\newcommand{\topline}{ %
    \arrayrulecolor{rulecolor}\specialrule{0.1em}{\abovetopsep}{0pt}%
    \arrayrulecolor{rulecolor}\specialrule{\lightrulewidth}{0pt}{0pt}%
    \arrayrulecolor{tableheadcolor}\specialrule{\aboverulesep}{0pt}{0pt}%
    \arrayrulecolor{rulecolor}
    }
\newcommand{\midtopline}{
    \arrayrulecolor{tableheadcolor}\specialrule{\aboverulesep}{0pt}{0pt}%
    \arrayrulecolor{rulecolor}\specialrule{\lightrulewidth}{0pt}{0pt}%
    \arrayrulecolor{white}\specialrule{\aboverulesep}{0pt}{0pt}%
    \arrayrulecolor{rulecolor}}
    \newcommand{\bottomline}{
    \arrayrulecolor{tableheadcolor}\specialrule{\aboverulesep}{0pt}{0pt}
    \arrayrulecolor{rulecolor}
    \specialrule{\heavyrulewidth}{0pt}{\belowbottomsep}
    \arrayrulecolor{rulecolor}\specialrule{\lightrulewidth}{0pt}{0pt}
    }
\newcolumntype{?}{!{\vrule width 1.4pt}}
\resizebox{\textwidth}{!}{%
\begin{tabular}{l|c|cccc?c|cccc}
\topline
\textbf{Method} &
\textbf{Backbone} &
  \textbf{AUROC}  &
  \textbf{AUC-PR} &
  \textbf{F1} &
  \textbf{Acc} &
  \textbf{Backbone} &
  \textbf{AUROC} &
  \textbf{AUC-PR} &
  \textbf{F1} &
  \textbf{Acc}  
\\ 
\midtopline \myrowcolour
\multicolumn{11}{c}{\textbf{\texttt{Wikibio-GPT3}}} \\ \midtopline
{\priorconf} & \multirow{6}{*}{\makecell{\texttt{gpt-3.5}\\ \texttt{-turbo}}} & 73.1 & 85.7 & 84.5 & 76.3 &  \multirow{6}{*}{\makecell{\texttt{Llama3-8B}\\\texttt{Instruct}}} & 71.0 & 85.7 & 85.5 & 75.9 \\
{\chainofthought} & & 71.3 & 83.4 & 85.2 & 76.4 &   & 72.0 & 84.4 & 85.3 & 75.7  \\
{\semantic} &  & 70.8 & 86.1 & 84.3 & 73.7 & & 60.3 & 78.7 & 84.5 & 73.6 \\
{\selfcheck} & & \textbf{82.6} & \textbf{91.3} & 86.6 & 80.0 & & \textbf{77.0} & \textbf{86.8} & 86.1 & 76.8 \\
{\focus} & & - & - & - & - & & 73.0 & 85.1 & \textbf{86.2} & 77.3 \\
\gr \textbf{{\alg}} & & 80.7  & 90.4 & \textbf{87.6} & \textbf{80.4} & & 74.0 & 86.2 & 86.1 & \textbf{77.5} \\
\midtopline \myrowcolour
\multicolumn{11}{c}{\textbf{\texttt{FELM-Science}}} \\ \midtopline
{\priorconf} & \multirow{7}{*}{\makecell{\texttt{gpt-3.5}\\ \texttt{-turbo}}} & 75.5 & 37.2 & 42.1 & 81.6 & \multirow{7}{*}{\makecell{\texttt{Llama3-8B}\\\texttt{Instruct}}} & 76.1 & 38.2 & 44.0 & 80.6 \\
{\chainofthought} & & 56.3 & 19.0 & 28.6 & 72.5 & & 61.6 & 25.7 & 32.6 & 71.2 \\
{\semantic} & & 59.1 & 25.6 & 32.6 & 75.2 & & 52.5 & 17.2 & 27.9 & 79.6  \\
{\selfcheck} & & 77.4 & 45.7 & 51.3 & \textbf{84.4} & & \textbf{77.8} & 39.5 & 49.1 & 76.5 \\
{\maieutic}  & & - & - & 27.2 & 82.6 & & - & - & 22.9 & 79.2 \\
{\focus} & & - & - & - & - & & 69.4 & 43.5 & 41.7 & 76.5 \\
\gr \textbf{{\alg}} &  & \textbf{79.1} & \textbf{52.3} & \textbf{56.5} & 81.6 & & \textbf{77.8} & \textbf{48.2} & \textbf{51.3} & \textbf{82.8} \\
\midtopline \myrowcolour
\multicolumn{11}{c}{\textbf{\texttt{FactCheckGPT}}} \\ \midtopline
{\priorconf} & \multirow{6}{*}{\makecell{\texttt{gpt-3.5}\\ \texttt{-turbo}}} & 76.0 & 52.6 & 53.6 & 71.5 & \multirow{6}{*}{\makecell{\texttt{Llama3-8B}\\\texttt{Instruct}}} & 71.9 & 47.6 & 50.8 & 60.5\\
{\chainofthought} & & 66.5 & 38.9 & 47.7 & 66.4 & & 72.5 & 43.3 & 53.2 & 66.1 \\
{\semantic} & & 56.3 & 33.9 & 40.9 & 65.9 & & 57.6 & 34.1 & 40.8 & 49.9 \\
{\selfcheck} & & 74.9 & 49.7 & 54.4 & 74.0 & & 72.5 & 46.4 & 51.9 & 65.9 \\
{\maieutic} & & - & - & 35.0 & 72.2 & & -& - & 39.8 & 66.8 \\
\gr \textbf{{\alg}} & & \textbf{79.4} & \textbf{54.3} & \textbf{60.2} & \textbf{75.3} & & \textbf{73.9} & \textbf{49.1} & \textbf{55.3} & \textbf{72.6}  \\
\midtopline
\bottomline

\end{tabular}
}
\caption{Hallucination detection performance of different methods. We report AUROC, ROC-PR, F1 score, and detection accuracy(Acc) for all methods with two backbone models. The best results are highlighted in \textbf{bold}.}
\label{tab: exp_main}
\end{table*}

\paragraph{Evaluation settings and metrics} We conduct hallucination detection at sentence-level on \texttt{Wikibio-GPT3} and \texttt{FactCheckGPT} and segment-level on \texttt{FELM-Science} following the default settings. Our evaluation metrics include the area under the receiver operator characteristic curve (AUROC), area under the precision-recall curve (AUC-PR), F1 score, and detection accuracy.  As F1 score and detection accuracy evaluation require a decision threshold, we search for the optimal threshold to maximize the F1 score for each method and compute the two metrics.
The hallucinated examples are considered as positive instances following the default configurations of these datasets.
Moreover, \texttt{Wikibio-GPT3} and \texttt{FELM-Science} exhibit significant class imbalances, so the detection accuracy on these datasets serve merely as reference points.

\paragraph{Baselines} We include the following baselines for comparison. (1) {\priorconf} , which directly queries the model's confidence on the truthfulness of each sentence or segment. (2) {\chainofthought}, which prompts the model to first generate a reasoning process before deciding the truthfulness. We adopt the prompting method from the official \texttt{FELM}~\cite{chen2023felm} dataset, with slight modifications for sentence-level and segment-level hallucination detection.
(3) {\selfcheck}~\cite{manakul2023selfcheckgpt}, which samples additional responses from the model and use the inconsistency between each response and the target statement for hallucination detection. Among the multiple variants, we choose {\selfcheck}-prompt with the best performance for comparison.
(4) {\maieutic}~\cite{jung2022maieutic}, which builds a belief tree via backward-chaining and then infers the truth-value of the original statement that resolves the inconsistencies.
(5) {\semantic}~\cite{kuhnsemantic, farquhar2024detecting}, which estimate the model's predictive uncertainty on the generation for hallucination detection.
(6) {\focus}~\citep{zhang2023enhancing}, which quantify the predictive uncertainty of the LLM on given texts to detect
hallucinations. Note that we only include {\focus} when the backbone is \texttt{Llama3} since it requires full access to the LLM.

\paragraph{Implementation details} 
We evaluate our methods and baselines using  \texttt{GPT-3.5-turbo-0125} and \texttt{Llama-3-8B-Instruct}.
For our method, we set the maximum belief tree depth to $2$. We employ greedy decoding during belief tree construction and prior belief estimation. The exception is the statement correction strategy, where we sample $5$ corrected statements using temperature $0.7$. Additionally, since each statement in \texttt{FactCheckGPT} is manually processed to ensure it contains only one property or fact, we do not apply statement decomposition when build the belief tree for statements from \texttt{FactCheckGPT}. We use the first 120 examples in the \texttt{Wikibio-GPT3} dataset to estimate the emission probability in our method, and validate it on the remaining examples and two other datasets. More details are in Appendix~\ref{implement_detail}.

\subsection{Experiment Results}
\label{subsec: exp_res}
\paragraph{Overall comparison} We evaluate the effectiveness of {\alg} with the experiment results in Table~\ref{tab: exp_main}. We highlight the following observations. First, our method achieves the best performance on \texttt{FELM-Science} and \texttt{FactCheckGPT} datasets across different backbones, demonstrating the superiority of our method. {\alg} improves upon the best baselines by 3\% - 9\% on AUROC and ROC-PR. The only exception is the \texttt{Wikibio-GPT3} dataset, where the {\selfcheck} is more effective for detection hallucinated outputs in biographies generated by LLMs.
Second, compared to {\selfcheck} which leverages contradictions between the target statement and the sampled responses for hallucination detection, our method is more effective on detecting hallucinated responses related to scientific knowledge. Both \texttt{FELM-Science} and \texttt{FactCheckGPT} datasets contain a significant proportion of questions on scientific knowledge, and our method achieves the best performance on them.
Third, chain-of-thought prompting is less effective in hallucination detection, especially on the \texttt{FELM-Science} dataset. This finding aligns with the experimental results in the original \texttt{FELM} dataset paper. Our experiments show that the model tends to regard the input sentence as true most of the time, leading to sub-optimal performance.
The Maieutic-prompting method, which is originall designed to verify the correctness of statements related to commonsense reasoning, is less effective for hallucination detection. Even on the \texttt{FELM-Science} and \texttt{FactCheckGPT} datasets which contain many statements about scientific knowledge, it remains less effective compared to other methods. We also find that the Semantic Uncertainty method is more effective on detection hallucinations in biography generation. However, its effectiveness diminishes when applied to the other two datasets. In contrast, our method consistently delivers competitive performance across a variety of benchmarks.

\begin{figure}
    \centering
    \subfigure
    {          
        \includegraphics[width=.23\textwidth, height=!]{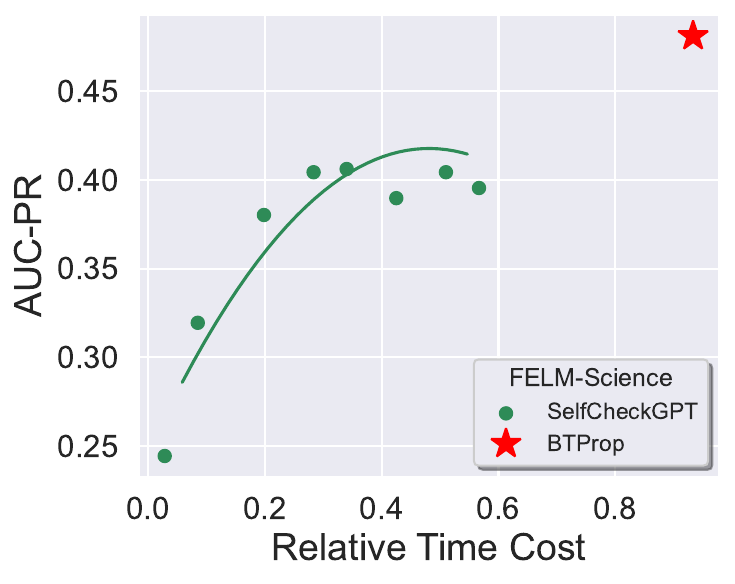}
    }
    \hspace{-3mm}
    \subfigure
    {          
        \includegraphics[width=.23\textwidth, height=!]{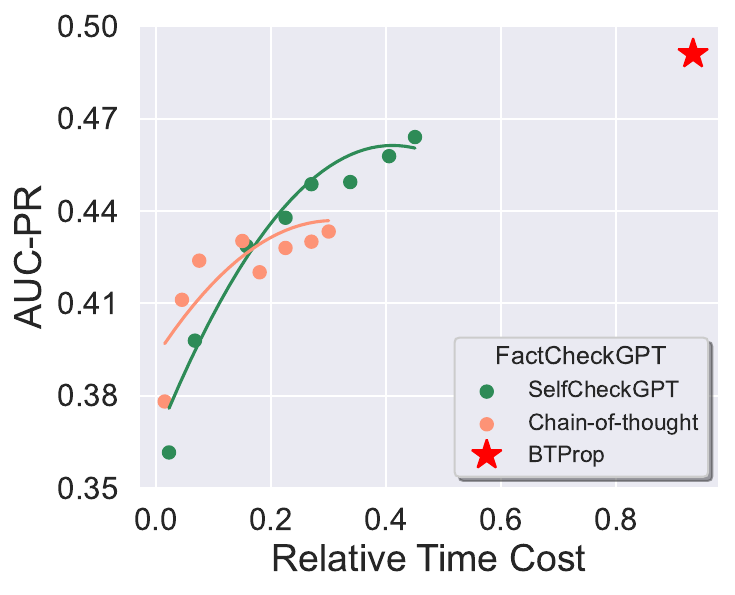}
    }
    \vspace{-6mm}
    \caption{Performance-efficiency comparison.} 
    \vspace{-3mm}
\label{fig: exp_tradeoff}
\end{figure}

\paragraph{Deployment efficiency}
One concern on our method is the belief tree construction would be inefficient. We compare the time cost of our method with the baselines on the \texttt{FELM-Science} and \texttt{FactCheckGPT} datasets, where our method significantly outperform baselines. 
We use the \texttt{gpt-3.5-turbo} as the backbone, and eliminate CoT prompting method on \texttt{FELM-Science} dataset due to its poor performance.
We visualize the trade-off between hallucination detection performance and the time cost in Figure~\ref{fig: exp_tradeoff}. The baseline methods, including {\selfcheck} and {\chainofthought} prompting, samples additional 20 responses for hallucination detection. We vary the number of samples and test the corresponding performance to visualize the trade-off. As depicted in the graph, the baseline performance shows diminishing returns in performance improvement with increased computational complexity. As time cost increases, the performance growth of the two baseline models gradually levels off, eventually reaching a point where additional time does not translate into significantly better performance. In comparison, our method can effectively trades off increased time cost for significantly improved performance. Despite its higher computational complexity, it achieves a marked improvement in performance metrics, underscoring its efficiency and effectiveness.

\paragraph{Hallucination detection in texts generated by the backbone model.}
Since the datasets in the above experiments consist of text generated by another model for hallucination detection, we also include texts generated by the backbone model itself to to provide a more comprehensive evaluation of our method’s effectiveness. Detailed results are available in Appendix~\ref{app: additional_exp_in_rebuttal}, where we show that our method can also effectively detect the hallucinations in the text generated by the backbone model itself.

\paragraph{Qualitative results} We visualize several examples of the constructed belief trees to demonstrate how the inconsistencies in model's beliefs help detect the hallucination. Due to space limit, we put these examples and the  analyses in Appendix~\ref{subsec: belief_tree_example}.

\begin{table}[]
    \centering
    \resizebox{0.95\linewidth}{!}{
    \begin{tabular}{l|cccc}
    \toprule[1pt]
     & AUROC & AUC-PR & F1 & Acc \\
    \midrule
    {\alg} & 79.1 & 52.3 & 56.5 & 81.6  \\
    \ \ \ \ Decomposition only & 67.9 & 31.2 & 43.3 & 79.6  \\
    \ \ \ \ Premise only & 70.2 & 30.7 & 40.4 & 78.6  \\
    \ \ \ \ Correction only & 77.4 & 47.5 & 53.0 & 81.5  \\
    {\priorconf} & 75.5 & 37.2 & 42.1 & 81.6  \\
    \bottomrule[1pt]
    \end{tabular}
    }
    \caption{Performance of {\alg} given different child node generation strategies.}
    \label{tab: ablation}
    \vspace{-3mm}
\end{table}

\paragraph{Ablation study: belief tree construction}
We introduce three strategies to build the belief tree: statement decomposition, generating supportive and contradictory premises, and statement correction. We perform ablation study to demonstrate the necessity to introduce these different strategies in Table~\ref{tab: ablation} using \texttt{gpt-3.5-turbo-0125} as the backbone model. Specifically, we evaluate the performance of our method when only use one child node generation strategies on the \texttt{FELM-Science} dataset. 
We highlight the following observations. First, only applying one child node generation strategy is sub-optimal. There exist a significant gap between using one strategy and using them all. Second, statement correction tends to contribute most to the performance improvement, but when combining it with other strategies, the performance can still be further improved.

\section{Conclusion}
\label{sec: conclusion}
In this paper, we propose {\alg}, a method that leverages the model's intrinsic capability for hallucination detection. Given a statement from the LLM-generated texts, our method organizes the model's intrinsic beliefs on neighboring statements in a belief tree.
By introducing the hidden Markov tree model, we convert the the hallucination detection into a posterior probability estimation problem and propose corresponding solutions to solve it. Experiment results have demonstrate the effectiveness of our method. The future direction would be how to further improve the belief tree construction method to make it more effective and efficient.

\section{Societal Impact and limitations}
In this paper, our primary goal is to develop an algorithm that can integrate the model's internal beliefs on logically-connected statements to detect hallucinations in texts generated by the LLM. Our method is designed to improve the trustworthiness of LLMs and make fully use of their intrinsic capabilities. Therefore, our method is less likely to introduce the unintended risks. We also assess the experiments to ensure they are devoid of any harmful content or adverse impacts.

While {\alg} improves the hallucination detection performance and enhance the trustworthiness of LLMs, it involves collecting a set of augmented statements (the belief tree) to perform inference. Therefore, the main limitation of our method is its high time cost. Constructing a belief tree necessitates multiple queries to the large language model, leading to significant delays. Additionally, the time complexity of generating child nodes increases exponentially as more layers are added to the tree. A potential solution to this issue could be to develop a mechanism that selectively expands nodes within the belief tree, thereby optimizing the process. 

\section*{Acknowledgments}
The work of Bairu Hou and Shiyu Chang was partially supported by National Science Foundation (NSF) Grant IIS-2338252, NSF Grant IIS-2207052, and NSF Grant IIS-2302730. The work of Jacob Andreas is supported by a Sloan Fellowship and the NSF under grant IIS-2238240. The computing resources used in this work were partially supported by the Accelerate Foundation Models Research program of Microsoft and CAIS Compute Cluster of Center for AI Safety.
\bibliography{custom}

\appendix

\section{Appendix / supplemental material}
\subsection{Implementation Details}
\label{implement_detail}
\paragraph{Data preprocessing} We preprocess the data in \texttt{Wikibio-GPT3}~\cite{manakul2023selfcheckgpt} and \texttt{FELM-Science}~\cite{chen2023felm} before evaluating our method and baselines. The data preprocessing includes two steps. First, we perform \emph{decontextualization} since the sentences in a response generated by the LLM might be context-dependent and contain pronouns or noun phrases that cannot be understood without full context. Take the first data instance from \texttt{Wikibio-GPT3} for demonstration, which is a biography of ``John Russell Reynolds'' generated by an LLM. The second sentence within the biography requiring hallucination detection is ``He was born in London, the son of a barrister, and was educated at Eton College and Trinity College, Cambridge.'' Without the context, we cannot identify the truthfulness of this sentence due to the pronoun. Therefore, we first decontextualize the sentences in the two datasets by prompting \texttt{gpt-3.5-turbo-0125}. The prompt is available in Figure~\ref{prompt: preprocessing}.

Second, we further manually clean up the dataset by filtering out some sentences that are not check-worthy but still annotated as ``true'' in the datasets. For example, the \texttt{FELM} contains some sentences such as ``Sure!'' and ``If you have any further questions or concerns, please let me know.''. These sentences are annotated as ``true'' and will be counted into performance evaluation. Since both two datasets contains not too many examples, we manually filter out these sentences to exclude them in our evaluation. The datasets after our preprocessing and filtering are available in the supplemental material and will be made open-sourced.

\paragraph{Implementation of our method} We use the vLLM~\cite{kwon2023efficient} to perform the inference of \texttt{Llama-3-8B-Instruct}. We use a single NVIDIA A100 80GB PCIe GPU to evaluate the performance and report the time cost in Figure~\ref{fig: exp_tradeoff}. For the emission probability estimation (Table~\ref{tab: emission probability}), we use the first 120 examples in the \texttt{Wikibio-GPT3} dataset (50\% of it) to compute the empirical estimation of the emission probability. The model we use is \texttt{gpt-3.5-turbo-0125}. Then we transfer the estimated emission probabilities to the remaining examples in \texttt{Wikibio-GPT3} and other datasets for performance evaluation. For \texttt{Llama-3-8B-Instruction}, we also use the same estimated emission probabilities. 

For the emission probability, we use first 120 examples of the \texttt{Wikibio-GPT3} dataset as the validation set and estimate the emission probability on it using \texttt{gpt-3.5-turbo-0125}. The estimated emission probabilities are in Table~\ref{tab: emission probability} and will used on the other testing examples of \texttt{Wikibio-GPT3} and other two datasets. 
Additionally, since we set the model confidence on child nodes generated by statement correction as 1.0 rather than probe the model's confidence, we find setting an individual set of emission probabilities for those statements improves the performance. Therefore, we tune the emission probabilities for child nodes generated by statement correction on the validation set. The emission probabilities for those child nodes we use are $p(S_u=1\mid Z_u = 1) = 0.8$ and $p(S_u=1\mid Z_u = 0) = 0.2$. During the belief tree construction process, we add several constraints to boost the efficiency. Specifically, we only apply statement decomposition to the root node, assuming the child nodes generated by the LLM is atomic statements that only contain one aspect of information. Also, we do not expand nodes generated by statement correction, as they are homogeneous to their parent node. 

\paragraph{Implementation of baselines} We follow the default configuration of each baselines. For {\selfcheck}, we sample 20 additional responses for hallucination detection. Similarly, we also sample 20 different answers for chain-of-thought prompting and aggregate them using self-consistency~\cite{wang2022self}. For Maieutic-Prompting, we use their prompts for the CREAK datasets to generate belief trees in our evaluation. For all methods, we use the \texttt{scikit-learn} package to compute the evaluation metrics including AUROC, AUC-PR, and F1 score.

\subsection{Additional Results}
\label{app: additional_exp_in_rebuttal}
\begin{table}[]
    \centering
    \resizebox{0.95\linewidth}{!}{
    \begin{tabular}{l|cccc}
    \toprule[1pt]
    \midrule
     & AUROC & AUC-PR & F1 & Acc \\
    \midrule
    Prior Confidence & 67.7 & 63.2 & 71.6 & 68.1  \\
    MSP	 & 70.5 & 73.3 & 69.3 & 57.4  \\
    Perplexity & 66.0 & 63.4 & 70.0 & 59.3  \\
    CCP & 75.8 & 78.1 & 72.4 & 69.5  \\
    {\alg} & 76.5 & 75.5 & 73.3 & 70.0  \\
    \midrule 
    \bottomrule[1pt]
    \end{tabular}
    }
    \caption{Performance of {\alg} given different child node generation strategies.}
    \label{tab: rebuttal_exp}
    \vspace{-3mm}
\end{table}

Since the datasets in the above experiments consist of text pre-generated by another model for hallucination detection, we also include texts generated by the hallucination detection model itself to further evaluate the effectiveness of our method. Specifically, we first generate biographical data using the prompt from \texttt{Wikibio-GPT3} dataset with \textit{Llama-3-8b-instruct}, following the ``claim-level uncertainty quantification'' setup of LM-Polygraph~\citep{fadeeva2023lm}. Then we use the same model to detect the hallucinations in its generations, where the ground-truth label for each claim is determined by FactScore~\citep{min2023factscore}. We evaluate the performance of our method and additional baselines including CCP~\citep{fadeeva2024fact}, maximum sequence probability (MSP), and Perplexity. The ground-truth label for each claim is determined by FactScore. The results are summarized in Table~\ref{tab: rebuttal_exp}.

Similar to the observation in our main experiments, our method performs competitively against these state-of-the-art uncertainty quantification techniques. Specifically, our approach achieves the highest AUROC, F1 scores, and accuracy, demonstrating its ability to reliably detect hallucinations. Please note that our method is not originally designed for the “claim-level” hallucination detection, where the biography data are decomposed into atomic claims for fact-checking. Nevertheless, our method still achieves competitive performance.

\subsection{Examples of Belief Trees}
\label{subsec: belief_tree_example}

\begin{figure}
    \centering
    \includegraphics[width=\linewidth, height=!]{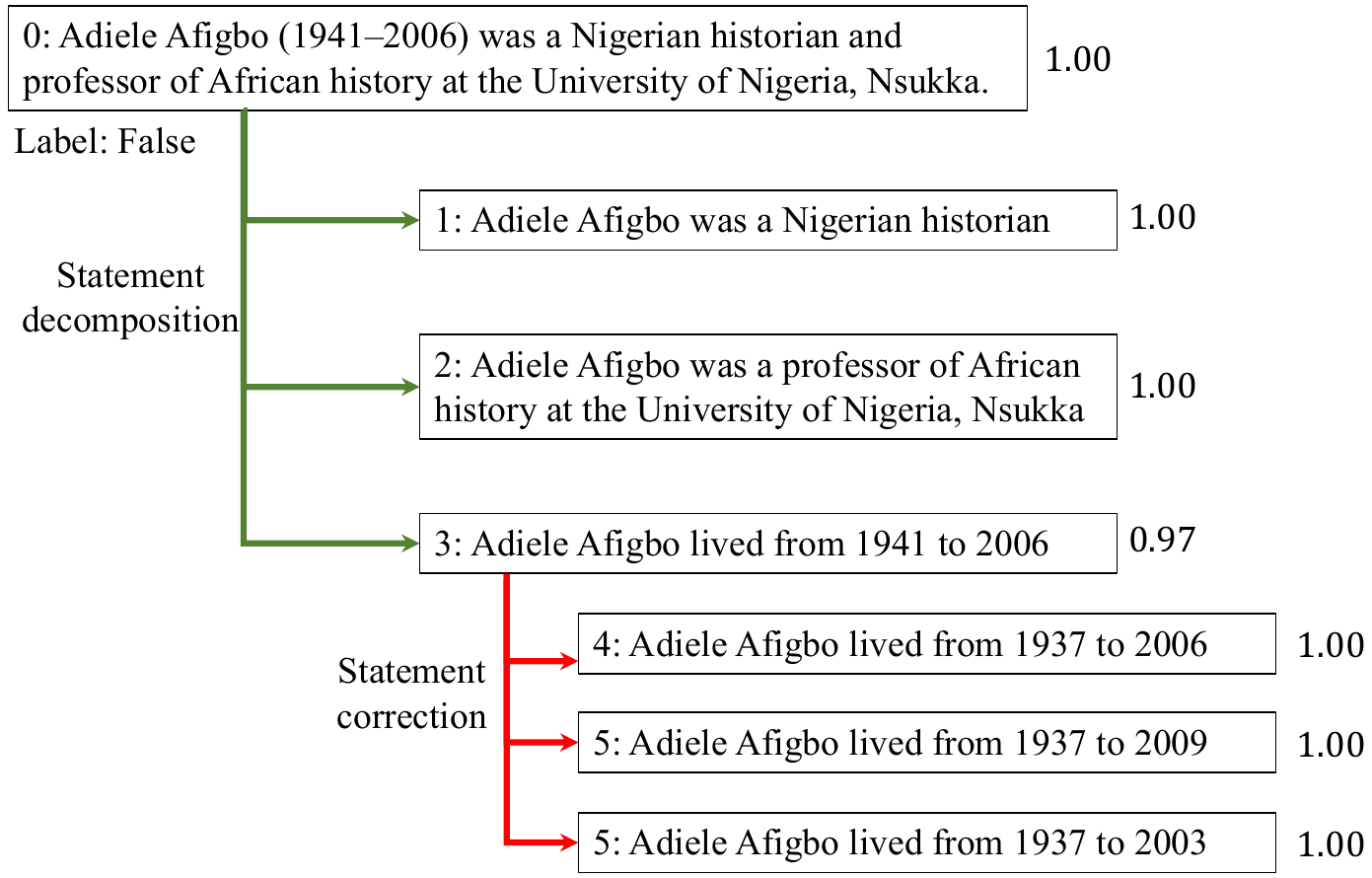}
    \caption{Belief tree example.} 
\label{fig: belief_tree_1}
\end{figure}
We visualize several constructed belief trees as well as how our method leverage the inconsistencies in model's beliefs for hallucination detection. We start with a simple example from the \texttt{Wikibio-GPT3} dataset~\cite{manakul2023selfcheckgpt} shown in Figure~\ref{fig: belief_tree_1}. Starting from the root node about \texttt{Adiele Afigbo}, the first layer contains 3 child nodes generated by statement decomposition. However, despite the statement is wrong, the model (\texttt{gpt-3.5-turbo-0125}) assign a high confidence score to both the original statement and the child nodes decomposed from the root node. Then, our method further generate child nodes for node 1,2, and 3. We display the child nodes of node 3 in the figure, which is generated by statement correction. With the three different statements about the date of death of that person, we decrease the model's confidence on node 3 and finally correct the model's belief on the root node. Note that in statement correction, we directly set the confidence of the generated child nodes as 1.0. For this example, if we query the model's confidence scores on node 4, 5, and 6, we will get confidence scores 0.97, 0.91, 0.88, respectively, which will still contradicts to their parent node.

\begin{figure}
    \centering
    \includegraphics[width=\linewidth, height=!]{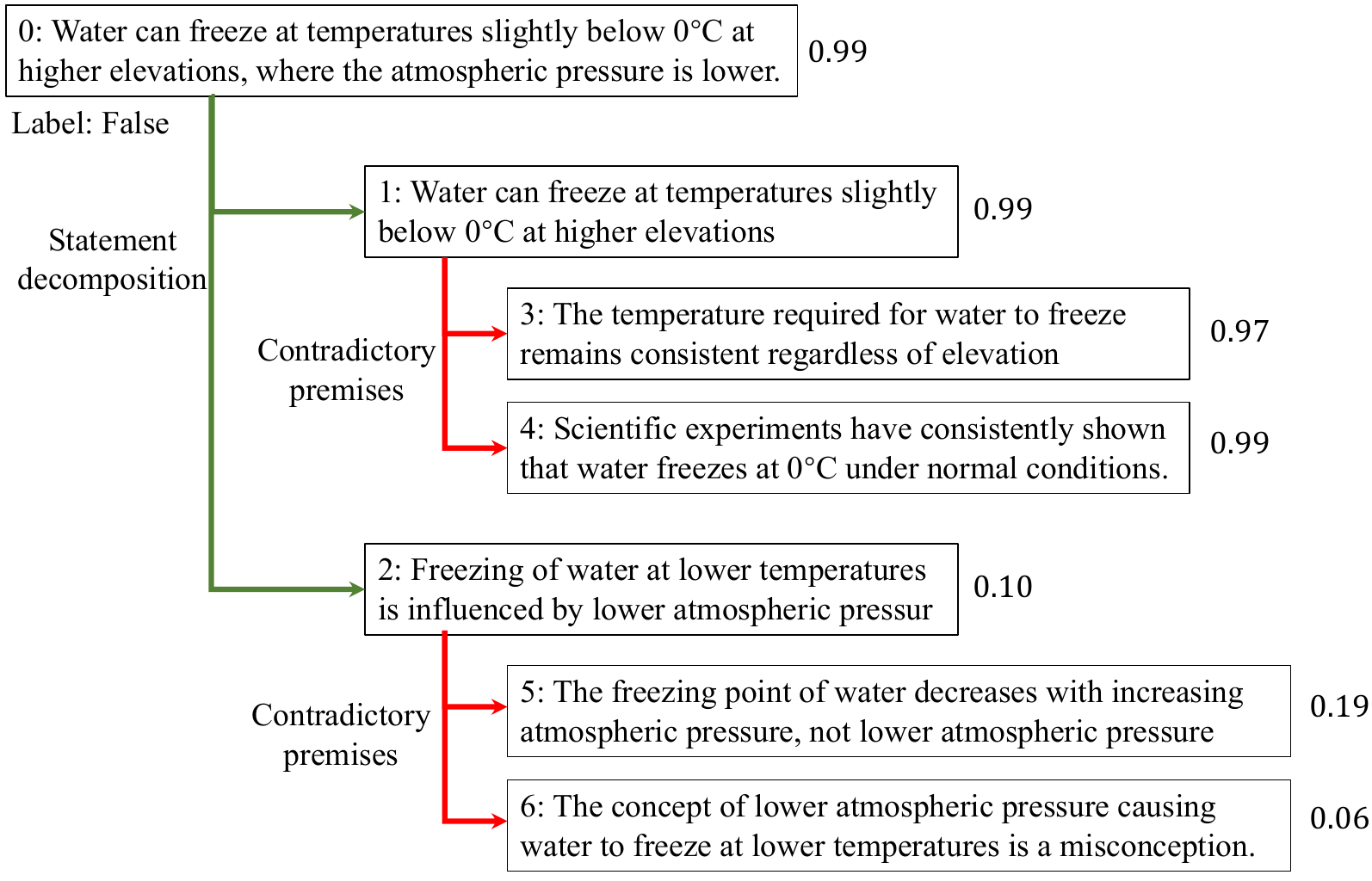}
    \caption{Belief tree example.} 
\label{fig: belief_tree_2}
\end{figure}
Another example of the belief tree is shown in Figure~\ref{fig: belief_tree_2}, which mainly consists of child nodes generated by supportive and contradictory premises. At the root node, the model assigns a high confidence score to the statement about freezing point of water from the \texttt{FELM} dataset~\cite{chen2023felm}. After the statement decomposition, the inconsistency is triggered due to model's low confidence on node 2. Furthermore, if we continuously generate premises for node 1, we get two additional child nodes (node 3 and 4) that are contradictory to their parent node. However, the model still assigns a high confidence to node 3 and 4. Within the belief propagation framework, the conditional probability of node 1 being true given the observations on node 3 and 4 will be decreased. Similarly, this effect will propagate to the root node and lead to a low posterior probability of the root node being true.

\begin{figure}
    \centering
    \includegraphics[width=\linewidth, height=!]{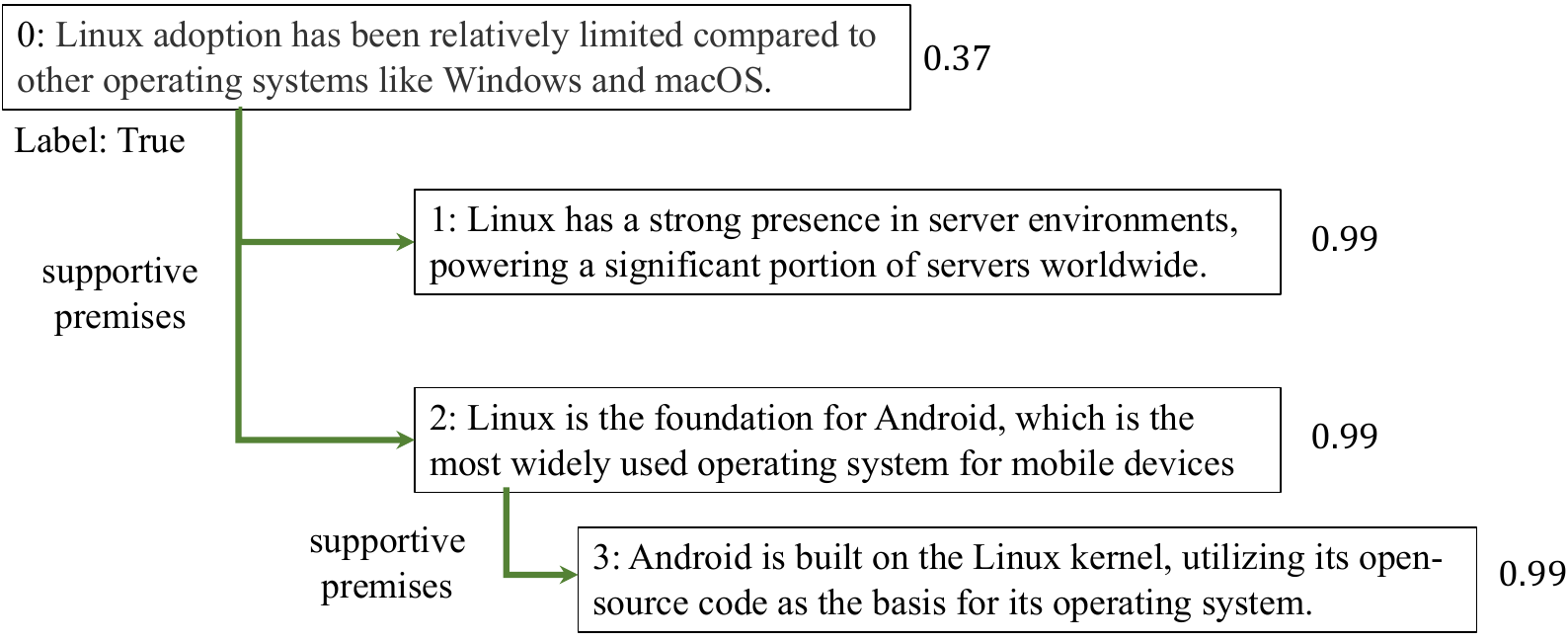}
    \caption{Belief tree example.} 
\label{fig: belief_tree_3}
\end{figure}
We then show one failure case of our method in Figure~\ref{fig: belief_tree_3}. The target statements comes from the \texttt{FactCheckGPT} dataset~\cite{wang2023factcheck}. Our method first generation two supportive premises with high confidence. These two child nodes are indeed reasonable and can support the opinion that  ``Linux operating system has a widespread adoption.'', thus contradicting to the root statement and further decreasing the model's confidence on the root node. We hypothesize that the original statement, ``Linux adoption has limited compared to other operating systems like Windows and macOS'' lacks sufficient specificity and could be interpreted from multiple perspectives. For example, it could refer to market share in personal computing versus cloud computing domains, making the ground-truth label nondeterministic.

\subsection{Inferring the Posterior Probability of $Z_0$}
\label{appendix: detailed_derivation}
The posterior probability $p(Z_{0} = T \mid \{S_u\})$ can be computed by:
\begin{equation*}
\small
    \begin{aligned}
        \frac{p(Z_0 = T, \{S_u\})}{p(Z_0 = T, \{S_u\}) + p(Z_0 = F, \{S_u\})},
    \end{aligned}
\end{equation*}
where $\{S_u\} = \{s_u\}$ refers to the set of all observed variables in the belief tree.

Therefore, the key of the inference on the belief tree is to compute the two joint probabilities $p(Z_0 = T, \{S_u\})$ and $p(S_1 = F, \bar{\bm X}_1 = \bar{\bm x}_1)$. 
Following the conditional independence assumption in the hidden Markov model, the truthfulness (\emph{i.e,}, states of hidden variables) of each node is determined by its parent node in the tree and the transition probability. Also, the model's confidence (\emph{i.e.}, states of observed variables) on each node is determined by the state of the corresponding variable and the emission probability. With such an assumption, they can be decomposed as:
\begin{equation}
\small
\label{eq: joint_probability}
    \begin{aligned}
        p(Z_0 = T, \{S_u\}) &= p(\{S_u\} \mid Z_0 = T) * p(Z_0 = T),
    \end{aligned}
\end{equation}
where $p(Z_0 = T)$ is the prior probability of the statement being true (note that we set it to be 0.5 in this paper). $p(\{S_u\} \mid Z_0 = T)$ is the conditional probability of all observed variables (exclude the root node) given the root node being true. Recall that we introduce a notation $\beta(z, u) = p(S_{\mathcal{T}(u)}|Z_u = z)$ to represent such conditional probabilities, where $S_{\mathcal{T}(u)}$ refers to the confidence scores (observed variables) of all nodes on the subtree rooted at node $u$. Therefore, the conditional probability $p(\{S_u\} \mid Z_0 = T)$ can be denoted as $\beta(T, 0)$ (node $0$, hidden variable $Z_0 = T$).

Without loss of generality, we discuss below how to compute $\beta(z, u)$ for an arbitrary node $u$ in the tree. The computation for $\beta(T, 0)$ and $\beta(F, 0)$ can be performed in exactly the same way. Specifically, we can further decompose $\beta(z, u)$ as:
\begin{equation}
\small
    \begin{aligned}
    \label{eq: independent_decompose}
        p(S_{\mathcal{T}(u)})\mid Z_u = z) & = p(S_u \mid Z_u = z)\cdot\\
        & \left\{\prod_{v\in \bm \gC(u)}p(S_{\mathcal{T}(v)}\mid Z_u = z)\right\}.
    \end{aligned}
\end{equation}
The probability $p(S_{\mathcal{T}(v)} \mid Z_u = z)$ can be decomposed as:
\begin{equation}
\small
    \begin{aligned}
    \centering
        & p(S_{\mathcal{T}(v)} \mid Z_u = z) \\
        &= \sum_{k \in \{T,F\}} p(S_{\mathcal{T}(v)} \mid Z_v = k) * p(Z_v = k \mid Z_u = z) \\
        &= \sum_{k \in \{T,F\}} \beta(v, k) * p(Z_v = k \mid Z_u = z).
    \end{aligned}
\end{equation}
Therefore, the conditional probability $\beta(z, u) = p(S_{\mathcal{T}(u)})\mid Z_u = z)$ can be computed in a recursive manner:
\begin{equation}
\small
\label{eq: beta_compute}
    \begin{aligned}
        &\beta(z, u) = \underbrace{p(S_u \mid Z_u = z)}_{\text{Emission Probability}}\cdot \\
        &\left\{\prod_{v\in \bm \gC(u)}\sum_{k \in \{T, F\}}\beta(v, k) * \underbrace{p(Z_v = k \mid Z_u = z)}_{\text{Transition Probability}} \right\}.
    \end{aligned}
\end{equation}
There are two types of probabilities in the above equation. First, $p(S_u \mid Z_u = z)$ is the ``emission probability" at node $u$. Second, $p(Z_v = k \mid Z_u = z)$ is the ``transition probability" from node $u$ to its child node $v$. The recursive computation in Equation~\ref{eq: beta_compute} will starts from the leaf nodes. When $v$ is the leaf node in the tree, $\beta(v, k)$ is actually $p(S_v  \mid Z_v = k)$, which is the emission probability at node $v$. Such a process propagates from bottom to up. Finally, we can compute $\beta(t,0)$ as well as the joint probability $p(Z_0 = T \mid \{S_u\})$ according to Equation~\ref{eq: joint_probability}.

\paragraph{Beyond conditional independence: transition probability for statement decomposition.}
The above computation of the posterior probability is based on the assumption of conditional independence. Assume the node $u$ have two child nodes, $v_1$ and $v_2$. Then we consider the transition probabilities from $u$ to $v_1$ and from $u$ to $v_2$ independently. However, this does not hold for the child nodes generated by statement decomposition, in which the truthfulness of the child nodes are influenced by their parent node simultaneously. If $u$ is true, then we know that both two child nodes are also true. In contrast, if $u$ is false, we can only infer that at least one of the child nodes are false. To handle such a case, we revise the probability computation in Equation~\ref{eq: independent_decompose} accordingly. Assume node $u$ has $m$ child nodes. The probability $\beta(z, u) = p(S_{\mathcal{T}(u)}) \mid Z_u = z)$ can be computed as:
\begin{equation*}
\small
    \begin{aligned}
    \centering
    & p(S_{\mathcal{T}(u)})\mid Z_u = z) = p(S_u \mid Z_u = z) \tilde{\beta}(u, z), \text{where} \\
    &\tilde{\beta}(u, z) = \sum_{Z_{\mathcal{C}(u)} \in \{T, F\}^m}p(Z_{\mathcal{C}(u)} \mid Z_u = z)  \prod_{v \in \mathcal{C}(u)} \beta(Z_v, v) \\
\end{aligned}
\end{equation*}
which is the Equation~\ref{eq:ud} in the main paper.

\subsection{Summary of the Algorithm}
\label{subsec: algrithm_table}
\begin{figure}
\vspace{-3mm}
    \begin{minipage}{\linewidth}
      \begin{algorithm}[H]
      \small
        \caption{\small {\alg} Algorithm}
\label{alg: adaboost}
        \begin{algorithmic}[1]
        \State \textbf{Input:} Statement $u_0$, maximum tree depth $d_{\mathrm{max}}$
        \State \textbf{Output:} The posterior probability $p(Z_0 = 1\mid S_{\mathcal{T}(0)})$.

        \State
        
        \State Initialize the belief tree $\mathcal{T}$ with root node $u_0$
        \State Initialize the leaf node set $\mathcal{N} = \{u_0\}$
        \While{$\mathcal{N} \ne \oslash$}   \Comment{Belief tree construction}
            \State Pop an element $u$ from $\mathcal{N}$
            \State Generate child nodes $\mathcal{C}(u)$ for $u$
            \For{Node $v \in \mathcal{C}(u)$}
                \State Add $v$ to $\mathcal{T}$
                \State Add $v$ to $\mathcal{N}$ if $d_{u}< d_{\mathrm{max}}$
            \EndFor
            
        \EndWhile
        
        \State
        \Function{GetBeta}{$u$, $z$}  \Comment{Compute $\beta(z,u)$ in Eq.~\ref{eq:ud}}
            \If{$\mathcal{C}(u) = \oslash$}
                \State\Return $p(s_u \mid Z_u = z)$
            \EndIf
            \For{$v \in \mathcal{C}(u)$}
                \State $\beta(v, 0) = \textsc{GetBeta}(v, 0)$
                \State $\beta(v, 1) = \textsc{GetBeta}(v, 1)$
            \EndFor
            \State Compute $\beta(u, z)$ according to Eq.~\ref{eq:ud}
        \State\Return $\beta(u, z)$
        \EndFunction
        \State
        \State $\beta(1, 0)$ = \textsc{GetBeta}($u = 0$, $z = 1$)
        \State $\beta(0, 0)$ = \textsc{GetBeta}($u = 0$, $z = 0$)
        \State $p(Z_0 = 1\mid S_{\mathcal{T}(0)}) = \beta(1,0) / (\beta(1,0) + \beta(0,0))$

        \end{algorithmic}
        \label{fig: algorithm_table}
      \end{algorithm}
    \end{minipage}
\end{figure}
We summarize the pipeline of our method in Algorithm~\ref{fig: algorithm_table}. The root node of the belief tree is the given statement $u_0$. During the belief tree construction process, we maintain a set $\mathcal{N}$ that contains all current leaf nodes of the belief tree. We recursively expand each leaf node $u \in \mathcal{N}$ by its child nodes $\mathcal{C}(u)$ and add these new child nodes to the belief tree if they are logically-connected to the current leaf node. These generated child nodes then become new leaf nodes. They will also be added to $\mathcal{N}$ and be expanded until the maximum depth is reached. Given the constructed belief tree and the emission and transition probabilities estimated from a held-out dataset, the posterior probability $\beta(Z_0 = 1\mid S_{\mathcal{T}(0)})$ can be then computed in a recursive manner, which will be further used to predict the truthfulness of the given statement.

\subsection{Submission Checklist}
\label{appendix: license}
\texttt{FactCheckGPT} is under Apache-2.0 license. \texttt{FELM} is under CC-BY-NC-SA-4.0 license, and \texttt{Wikibio-GPT3} is under CC-BY-SA-3.0 license. We have cited them accordingly in the main paper.
All of these  datasets are in English.
These datasets are created to evaluate the hallucination detection performance, and our usages are consistent with their intended use.
We also manually check the datasets and confirm that they do not contains any information that names or uniquely identifies individual people or offensive content.

We leverage ChatGPT to help develop some of the prompt we use in the experiments and improve the writing.

\subsection{Prompt}
\label{appendix: prompt}
In this section, we list all the prompt used in this paper, including belief tree construction, prior confidence estimation, and data preprocessing.

\paragraph{Belief tree construction}
Given a statement from the LLM output, we use the following prompt \texttt{gpt-3.5-turbo-0125} to perform \emph{statement decomposition}, which is shown in Figure~\ref{prompt: decompose_chatgpt}. In the instruction, we specify the requirements to extract check-worthy claims and provide the model with several examples. We also include an additional special example, where the given sentence is actually a subjective opinion, to prevent the model from decomposing sentences that are actually not check-worthy. We use a similar prompt for \texttt{Llama-3-8b-Instruction} as shown in Figure~\ref{prompt: llama3_decompose}.

To generate \emph{supportive and contradictory premises}, we prompt the model as follows: If the model judge the given statement as true, then it needs to generate several explanations to its judgment, which form the supportive premises. In contrast, if the model believes the given statement is false, then it will generate explanations to its judgment, which form the contradictory premises. The prompts we used are listed in Figure~\ref{prompt: supportive} and Figure~\ref{prompt: contradictory}. We first prompt the model with the prompt in Figure~\ref{prompt: supportive}. If the model judges the statement as true and generates supportive premises, then these premises will be returned and contradictory premises will not be generated. If the model judges the statement as false, then we prompt the model with the prompt in Figure~\ref{prompt: contradictory} for contradictory premises. Although there might be better prompting strategies to generate such premises, finding the optimal prompts and prompting strategies is out of the scope of this paper. Therefore, we leave it for future work.

Finally, to generate child nodes via \emph{statement correction}, we adopt the following pipeline. First, we prompt the model using prompts listed in Figure~\ref{prompt: correction_chatgpt_1step} and Figure~\ref{prompt: correction_llama_1step} to generate a question about the key pieces of information in the statement. After that, we feed the generated question to the LLM again without any other prompt to get its answer. Finally, we ask the model to revise the original statement according to the ``background knowledge", which is the answer generated by itself. By doing this, we expect to better utilize the inconsistency across model's beliefs for child node generation. The prompt for the last step is shown in Figure~\ref{prompt: correction_3step}

To select the most appropriate strategies for each node, we use the following prompts to ask the LLM to output the most suitable strategies, which is displayed in Figure~\ref{prompt: routing_chatgpt} and Figure~\ref{prompt: routing_llama}.

\paragraph{Prior confidence estimation} When prompt \texttt{gpt-3.5-turbo-0125} for the confidence score on the truthfulness of a statement, we use the following prompt: \texttt{True or False? \{target\_statement\}}. For Llama-3 model, we find it sometimes refuse to judge the truthfulness and simply output it requires additional context. To enforce the model to output the confidence score, we change the prompt accordingly:

\begin{figure*}[h]
\begin{tcolorbox}
\begin{lstlisting}[style=text]
Are the following statements true or false? For each of the following statements, determine whether it is true or false. Provide a response of 'True' if the statement is correct, or 'False' if the statement is incorrect.

Remember:
    - You **do not** have access to additional information or external data.
    - If verifying the statement requires such external data or context, regard it as false.
    - Directly output "True" or "False" without adding any markers.
\end{lstlisting}
\end{tcolorbox}
\caption{Prompt for confidence estimation (\texttt{Llama-3-8B-Instruct}).}
\end{figure*}

\begin{figure*}[h]
\begin{tcolorbox}
\begin{lstlisting}[style=text]
**Rewrite Texts for Clarity**

In this task, you will receive one paragraph and one target statement extracted from it. The target statement is context-dependent, which makes the statement hard for us to understand without context and check its truthfulness. Therefore, your task is to rewrite the statement to reduce context dependency. Specifically,
   - Pronoun resolution: Replacing pronouns like "this," "the," "that," "he," "she," and "they" with specific nouns or names they refer to in the original paragraph. You should always use the full names.
   - If the target statement only use the first/last name to refer to the main entity, replace the first/last name with the full name of the entity if available.

Note: do not modify the semantics of the sentence. Do not add new information or your own descriptions into the statement.

**Input/Output Format**
The input will be provided with the format as below:
Original paragraph: <the original text>

Target statement: <the target statement needing rewrite>

Format your output as:
Output: <the target statement after rewrite>
\end{lstlisting}
\end{tcolorbox}
\caption{Prompt for data preprocessing.}
\label{prompt: preprocessing}
\end{figure*}

\begin{figure*}[h]
\begin{tcolorbox}
\begin{lstlisting}[style=text]
**Fact-Checking Claims Extraction:**

**Objective:** Analyze the provided statement to extract and list all distinct factual claims that require verification. Each listed claim should be verifiable and not overlap in content with other claims listed.

**Instructions:**
1. **Identify Factual Claims**:
   - Identify parts of the statement that assert specific, verifiable facts, including:
      - Statistical data and measurements.
      - Historical dates and events or other information.
      - Direct assertions about real-world phenomena, entities, events, statistics
      - Conceptual understandings and theories.
   - In every claim, alway use the **full names** when referring to any concept, person, or entity. **Avoiding the use of pronouns or indirect references** that require contextual understanding.

2. List each verifiable claim separately. Ensure that each claim is distinct and there is no overlap in the factual content between any two claims.
   - If a single claim is repeated in different words, list it only once to avoid redundancy.

3. **Output:**
   - If there are multiple check-worthy claims, list each one clearly and separately.
   - If there is only one check-worthy claim, output just that one claim.
   - If no part of the statement contains verifiable facts (e.g., purely subjective opinions, hypothetical scenarios), output the following message: "Claim 1: No check-worthy claims available."

**Output Format**:
Your output should be organized as follows:
Claim 1: <the first claim>
Claim 2: <the second claim>
Claim 3: <the third claim (if necessary)>
...

**Examples:**
Statement: According to recent data, China has surpassed the United States in terms of GDP when measured using Purchasing Power Parity (PPP), and India is projected to overtake China by 2030.
Claim 1: China has surpassed the United States in terms of GDP when measured using Purchasing Power Parity (PPP).
Claim 2: India is projected to overtake China in terms of GDP by 2030."

Statement: The world's largest desert is Antarctica, and it is larger than the Sahara.
Claim 1: The world's largest desert is Antarctica.
Claim 2: Antarctica is larger than the Sahara.

Statement: I think pizza is the best food ever!
Claim 1: No check-worthy claims available.

Statement: The software 'Photoshop' was released by Adobe Systems in 1988.
Claim 1: The software 'Photoshop' was released by Adobe Systems in 1988.
\end{lstlisting}
\end{tcolorbox}
\caption{The prompt for statement decomposition using \texttt{gpt-3.5-turbo-0125}.}
\label{prompt: decompose_chatgpt}
\end{figure*}

\begin{figure*}[h]
\begin{tcolorbox}
\begin{lstlisting}[style=text]
**Fact-Checking Claims Extraction:**

**Objective:** Analyze the provided statement to extract and list all distinct factual claims that require verification. Each listed claim should be verifiable and not overlap in content with other claims listed.

**Instructions:**
1. Identify parts of the statement that assert specific, verifiable facts. In every claim, alway use the **full names** when referring to any concept, person, or entity. **Avoiding the use of pronouns or indirect references** that require contextual understanding.

2. List each verifiable claim separately. Ensure that each claim is distinct and there is no overlap in the factual content between any two claims.

3. Exclude any statements that are purely hypothetical, assume theoretical scenarios, or are speculative in nature. These do not contain verifiable factual claims. For example, statements involving assumptions ("Let's assume..."), theoretical discussions ("consider whether..."), or purely speculative scenarios should not be considered as containing verifiable claims.

4. If the statement does not contain any verifiable facts, output the following message: "Claim 1: No check-worthy claims available."

**Output Format**:
Your output should be organized as follows:
Claim 1: <the first claim>
Claim 2: <the second claim>
Claim 3: <the third claim (if necessary)>
...

**Examples:**
Statement: According to recent data, China has surpassed the United States in terms of GDP when measured using Purchasing Power Parity (PPP), and India is projected to overtake China by 2030.
Claim 1: China has surpassed the United States in terms of GDP when measured using Purchasing Power Parity (PPP).
Claim 2: India is projected to overtake China in terms of GDP by 2030."

Statement: The world's largest desert is Antarctica, and it is larger than the Sahara.
Claim 1: The world's largest desert is Antarctica.
Claim 2: Antarctica is larger than the Sahara.

Statement: I think pizza is the best food ever!
Claim 1: No check-worthy claims available.

Statement: Let's assume there is a highest prime number and consider its implications on number theory.
Claim 1: No check-worthy claims available.
\end{lstlisting}
\end{tcolorbox}
\caption{The prompt for statement decomposition using \texttt{Llama-3-8b-Instruct}.}
\label{prompt: llama3_decompose}
\end{figure*}

\begin{figure*}[h]
\begin{tcolorbox}
\begin{lstlisting}[style=text]
**Finding Supportive Premises**

Is the following statement true or false? If it is true, list several supportive premises for it.

**Important Rules:**
1. Each premise should be clearly stated and directly relevant to the target statement. Avoid ambiguity and ensure that the connection to the target statement is evident
2. Do not use pronouns in generated premises. Ensure each premise can be understood clearly without any context. For each generated premise, you should always use the full name of each person, event, object, etc.

**Input/Output Format**:
Your output should be organized as follows.
Judgement: <True or False>
Premise 1: <the first premise>
Premise 2: <the second premise>
...

In contrast, if the statement is false, you simly output:
Judgement: False
Premise 1: No supportive premises applicable.

**Examples:**
Target statement: Renewable energy sources will lead to a decrease in global greenhouse gas emissions.
Judgement: True
Premise 1: Renewable energy sources produce electricity without emitting carbon dioxide.
Premise 2: Increasing the adoption of renewable energy reduces reliance on fossil fuels, which are the primary source of industrial carbon dioxide emissions.

Target statement: Eating carrots improves night vision.
Judgement: False
Premise 1: No supportive premises applicable.

Statement: Historical literacy enhances a society's ability to make informed decisions.
Judgement: True
Premise 1: Understanding historical events provides context for current issues, enabling citizens to make decisions that consider past outcomes and lessons.
Premise 2: Historical literacy fosters critical thinking skills, which are crucial in analyzing information and making reasoned decisions.
Premise 3: Societies with high levels of historical awareness can recognize and avoid the repetition of past mistakes.
\end{lstlisting}
\end{tcolorbox}
\caption{The prompt for generation of supportive premises.}
\label{prompt: supportive}
\end{figure*}

\begin{figure*}[h]
\begin{tcolorbox}
\begin{lstlisting}[style=text]
**Finding Contradictory Premises**

Is the following statement true or false? If it is false, list several contradictory premises for it.

**Important Rules:**
1. Each premise should be clearly stated and directly relevant to the target statement. Avoid ambiguity and ensure that the connection to the target statement is evident
2. Do not use pronouns in generated premises. Ensure each premise can be understood clearly without any context. For each generated premise, you should always use the full name of each person, event, object, etc.

**Input/Output Format**:
Your output should be organized as follows.
Judgement: <True or False>
Premise 1: <the first premise>
Premise 2: <the second premise>
...

In contrast, if the statement is false, you simply output:
Judgement: True
Premise 1: No contradictory premises applicable.

**Examples:**
Target statement: Renewable energy sources will lead to a decrease in global greenhouse gas emissions.
Judgement: True
Premise 1: No contradictory premises applicable.

Target statement: Eating carrots improves night vision.
Judgement: False
Premise 1: The belief that eating carrots improves night vision stems from World War II propaganda, not from scientific evidence.
Premise 2: While carrots are rich in vitamin A, which is necessary for maintaining healthy vision, they do not enhance night vision beyond normal levels.

Statement: The introduction of invasive species does not impact native biodiversity.
Judgement: False
Premise 1: Invasive species often compete with native species for resources, leading to a decline in native populations.
Premise 2: Studies show that invasive species can alter the natural habitats of native species, negatively affecting their survival rates.
Premise 3: The introduction of the invasive zebra mussel in North American waterways has led to significant declines in the populations of native mussels.
\end{lstlisting}
\end{tcolorbox}
\caption{The prompt for generation of contradictory premises.}
\label{prompt: contradictory}
\end{figure*}

\begin{figure*}[h]
\begin{tcolorbox}
\begin{lstlisting}[style=text]
Given the following claim, your tasks include:
    1. Identify the key pieces of information critical for fact-checking to determine its truthfulness. 
    2. Create a masked version of the claim by masking these key pieces of information
    3. Generate a question asking for the key pieces of information

Rules:
1. Do not mask the grammatical subject of the sentence -- the actor, entity, or object that performs the action in the sentence's main clause. Also, following the format of the below examples in your output.

**Examples:**
Statement: Bitcoin was created in 2009 by an anonymous entity known as Satoshi Nakamoto.
Masked statement: Bitcoin was created in 2009 by an anonymous entity known as [which person].
Question: Who created Bitcoin in 2009?

Statement: The iPhone was first released by Apple in 2007.
Masked statement: The iPhone was first released by Apple in [what year].
Question: Was the iPhone first released by Apple in 2007?

Statement: The speed of light in a vacuum is approximately 299,792 kilometers per second.
Masked statement: 'Romeo and Juliet' was written by Shakespeare in [what time period].
Question: What is the speed of light in a certain medium?

Statement: "Attention Is All You Need" is a paper written by Ashish Vaswani, Noam Shazeer, Niki Parmar, Jakob Uszkoreit, Llion Jones, Aidan N. Gomez, Lukasz Kaiser, Illia Polosukhin.
Masked statement: "Attention Is All You Need" is a paper written by [whom]
Question: Who was/were the authors of the paper "Attention Is All You Need"?

Statement: The Great Wall of China is visible from space.
Masked statement: The Great Wall of China is [visible or invisible] from space.
Question: Is the Great Wall of China visible from space?

Statement: The headquarters of the United Nations is located in New York City
Masked statement: The headquarters of the United Nations is located in [which city].
Question: Where is the headquarters of the United Nations located?
\end{lstlisting}
\end{tcolorbox}
\caption{Step 1 in statement correction: question generation. This prompt is used for \texttt{gpt-3.5-turbo-0125}.}
\label{prompt: correction_chatgpt_1step}
\end{figure*}

\begin{figure*}[h]
\begin{tcolorbox}
\begin{lstlisting}[style=text]
Given a statement, your task is to identify a general question that can be used to check the truthfulness of the statement. The question should directly address the claim to confirm or refute it without seeking additional detailed information.

**Examples:**
Statement: Bitcoin was created in 2009 by an anonymous entity known as Satoshi Nakamoto.
Question: Who created Bitcoin in 2009?

Statement: The iPhone was first released by Apple in 2007.
Question: When was iPhone first relased by Apple?

Statement: 'Romeo and Juliet' was written by Shakespeare in the late 16th century.
Question: When was 'Romeo and Juliet' was written by Shakespeare?

Statement: "Attention Is All You Need" is a paper written by Ashish Vaswani, Noam Shazeer, Niki Parmar, Jakob Uszkoreit, Llion Jones, Aidan N. Gomez, Lukasz Kaiser, Illia Polosukhin.
Question: Who was/were the authors of the paper "Attention Is All You Need"?

Statement: The Eiffel Tower is located in Paris
Question: Where is the Eiffel Tower located?

Statement: Albert Einstein was a physicist.
Question: What profession was Albert Einstein?
\end{lstlisting}
\end{tcolorbox}
\caption{Step 1 in statement correction: question generation. This prompt is used for \texttt{Llama-3-8b-Instruct}. We adjust the prompt since the masked statement is not used in the prompt for \texttt{gpt-3.5-turbo-0125}.}
\label{prompt: correction_llama_1step}
\end{figure*}

\begin{figure*}[h]
\begin{tcolorbox}
\begin{lstlisting}[style=text]
**Background Knowledge**: {the model answer in step 2}  

Leverage the above provided knowledge and your own knowledge to review the correctness of following statement:

**Statement**: {statement}

Instruction:
    - If the statement is correct, output it unchanged.
    - If the statement is **not mentioned in the background knowledge and its correctness cannot be determined**, you should also directly output the statement **unchanged**.
    - If the statement is wrong, revise only the parts of the statement that are incorrect, to align with the background knowledge. Do not add any additional sentences or details.

**Output Format:**
Format your output as:
Revised Answer: <Display the original statement if it is correct or not mentioned in the background knowledge; display the revised statement if it is inaccurate>
\end{lstlisting}
\end{tcolorbox}
\caption{Step 3 in statement correction: statement revision.}
\label{prompt: correction_3step}
\end{figure*}

\begin{figure*}[h]
\begin{tcolorbox}
\begin{lstlisting}[style=text]
**Task: Choose the Best Strategy for Premise Generation**

We need to generate several premises for a given target statement. These premises could either support or contradict the target statement. Particularly, we have 2 techniques for premise generation:

1. Logical Relationships: This involves creating premises based on entailment or contradiction. You can generate premises that either support or contradict the target statement. 

2. Statement Perturbation: Create variations of the statement by altering key details to form contradictory premises.

Given these techniques, your task is to select the most suitable technique given a particular statement. Follow the guidelines below to select the most suitable technique.

**Important Guidelines**:
1. Prioritize logical relationships. The logical relation strategy is broadly applicable, as long as it is straightforward to find supportive or contradictory premises.
2. If a statement contains particular names, numbers, timestamps, or other conditions that can be varied to generate contradictory premises, consider statement perturbation.
3. If both two methods are applicable, select them together and output "both".

**Output Format:**
Your selection should be one of "Logical Relation", "Statement Perturbation", and "both".

Target statement: Eating a balanced diet improves overall health.
Output: Logical Relation

Target statement: 'War and Peace' was a book written by Leo Tolstoy.
Output: Statement Perturbation

Target statement: Water boils at 100 degrees Celsius at sea level
Output: both

Target statement: Mount Everest is 8,848 meters tall.
Output: Statement Perturbation

Target statement: Artificial intelligence will replace most human jobs in the future
Output: Logical Relation
\end{lstlisting}
\end{tcolorbox}
\caption{Prompt for strategy selection (\texttt{gpt-3.5-turbo-0125}).}
\label{prompt: routing_chatgpt}
\end{figure*}

\begin{figure*}[h]
\begin{tcolorbox}
\begin{lstlisting}[style=text]
**Instruction for Choosing the Best Strategy for Premise Generation**

When tasked with generating premises for a given target statement, the choice of strategy-Logical Relationships or Statement Perturbation-should be determined based on the nature of the statement and the desired objectives. Use the following guidelines to route the choice:

### When to Use Logical Relationships
Opt for Logical Relationships when:
- **The Statement is Abstract or Principled**: Ideal for statements that explore broad principles, ethics, or abstract concepts. This method helps in drawing deep logical entailments or contradictions.
- **Complex Relationships or Conditions**: When the statement involves complex logical or conditional relationships, using this method clarifies or challenges these intricacies.
- **Requirement for Detailed Analysis**: For statements needing precise and formal argumentation, especially in academic or technical discussions.

### When to Use Statement Perturbation
Choose Statement Perturbation when:
- **The Statement is Specific and Concrete**: Best for statements with explicit details like scenarios, dates, locations, names, numbers or timestamps. Altering these elements generates varied premises.
- **Exploration of Counterfactuals or Hypotheticals**: Useful for creating imaginative or scenario-based premises by modifying key details to envision different outcomes.
- **Sensitivity to Detail Changes**: When minor modifications in the statement can significantly alter its implications or truth value.

**Output Format:**
Your selection should be one of "Logical Relation", "Statement Perturbation", and "both". Here "both" means both two methods are applicable.

**Examples:**
Target statement: Eating a balanced diet improves overall health.
Output: Logical Relation

Target statement: Countries with higher investment in education consistently rank higher in global innovation indexes
Output: Logical Relation

Target statement: 'War and Peace' was a book written by Leo Tolstoy.
Output: Statement Perturbation

Target statement: Water boils at 100 degrees Celsius at sea level
Output: both

Target statement: Mount Everest is 8,848 meters tall.
Output: Statement Perturbation

Target statement: The bridge will remain intact even if 75 cars drive on it simultaneously if the cars are lightweight
Output: Statement Perturbation

Target statement: Artificial intelligence will replace most human jobs in the future
Output: Logical Relation
\end{lstlisting}
\end{tcolorbox}
\caption{Prompt for strategy selection (\texttt{Llama-3-8b-Instruction}).}
\label{prompt: routing_llama}
\end{figure*}
\end{document}